\documentclass[lettersize,journal]{IEEEtran}
\usepackage{amsmath,amsfonts}

\usepackage{algorithmicx}
\usepackage{algorithm}
\usepackage{array}
\usepackage[caption=false,font=footnotesize]{subfig}
\usepackage{textcomp}
\usepackage{stfloats}
\usepackage{url}
\usepackage{verbatim}
\usepackage{graphicx}
\usepackage{cite}
\usepackage{multirow}
\usepackage{rotating}
\usepackage{booktabs}
\usepackage{algpseudocode}
\usepackage{wrapfig}
\usepackage{subfig}
\usepackage{tabularx}
\usepackage{bm}
\usepackage{enumerate}
\usepackage{color}
\usepackage{makecell}
\graphicspath{{images/}}
\begin{document}

\title{Information Recovery-Driven Deep Incomplete Multiview Clustering Network}

\author{Chengliang Liu,~\IEEEmembership{Graduate Student member,~IEEE,}
Jie Wen*,~\IEEEmembership{Member,~IEEE,}
Zhihao Wu,
Xiaoling Luo,\\
Chao Huang,
Yong Xu,~\IEEEmembership{Senior member,~IEEE}

\thanks{This work is partially supported by the Shenzhen Science and Technology Program under Grant No. RCBS20210609103709020, National Natural Science Foundation of China under Grant No. 62006059 and 62276079, Shenzhen Higher Education Stability Support Program Project under Grant No. GXWD20220811173317002, and CAAI-Huawei MindSpore Open Fund under Grant No. CAAIXSJLJJ-2022-011C.}
\thanks{Chengliang Liu, Jie Wen, Zhihao Wu, Xiaoling Luo, and Yong Xu are with the Shenzhen Key Laboratory of Visual Object Detection and Recognition, Harbin Institute of Technology, Shenzhen 518055, China (Email: liucl1996@163.com; jiewen\_pr@126.com; horatio\_ng@163.com; xiaolingluoo@outlook.com; yongxu@ymail.com).}
\thanks{Chao Huang is with the School of Cyber Science and Technology, Sun Yat-sen University (Shenzhen Campus), Shenzhen, China (Email: huangchao\_08@126.com).}
\thanks{Yong Xu is also with the Pengcheng Laboratory, Shenzhen 518055, China.}
\thanks{Corresponding author: Jie Wen (Email: jiewen\_pr@126.com).}
}
\markboth{Journal of \LaTeX\ Class Files,~Vol.~14, No.~8, August~2021}%
{Shell \MakeLowercase{\textit{et al.}}: A Sample Article Using IEEEtran.cls for IEEE Journals}


\maketitle

\begin{abstract}
Incomplete multi-view clustering is a hot and emerging topic. It is well known that unavoidable data incompleteness greatly weakens the effective information of multi-view data. To date, existing incomplete multi-view clustering methods usually bypass unavailable views according to prior missing information, which is considered as a second-best scheme based on evasion. Other methods that attempt to recover missing information are mostly applicable to specific two-view datasets. To handle these problems, in this paper, we propose an information recovery-driven deep incomplete multi-view clustering network, termed as RecFormer. Concretely, a two-stage autoencoder network with the self-attention structure is built to synchronously extract high-level semantic representations of multiple views and recover the missing data. Besides, we develop a recurrent graph reconstruction mechanism that cleverly leverages the restored views to promote the representation learning and the further data reconstruction. Visualization of recovery results are given and sufficient experimental results confirm that our RecFormer has obvious advantages over other top methods.
\end{abstract}


\section{Introduction}
\IEEEPARstart{I}{t} is well known that multi-view data depicts the observed objects from different perspectives \cite{tao2019marginalized,wen2023deep,liu2023dicnet,li2021consensus}. Compared with traditional single-view data, this heterogeneous data retains multi-level semantic information \cite{wen2022survey,LUO2021108104,wang2021fast,liang2021af}. In the past few years, multi-view clustering, as a novel representation learning method, has aroused extensive research enthusiasm and has been widely used in related fields such as data analysis \cite{hu2020dmib,liu2022localized,li2021incomplete}. However, conventional multi-view clustering methods usually assume that obtaining complete multi-view data is a matter of course, which obviously goes against practical experience. Therefore, a number of incomplete multi-view clustering (IMC) approaches have been developed to fit the increasingly common incomplete multi-view data \cite{ZhuMa2022,li2022refining,li2022parameter}, which is also the focus of our paper.
\begin{figure}[t!]
	\begin{center}
		\includegraphics[width=0.7\linewidth]{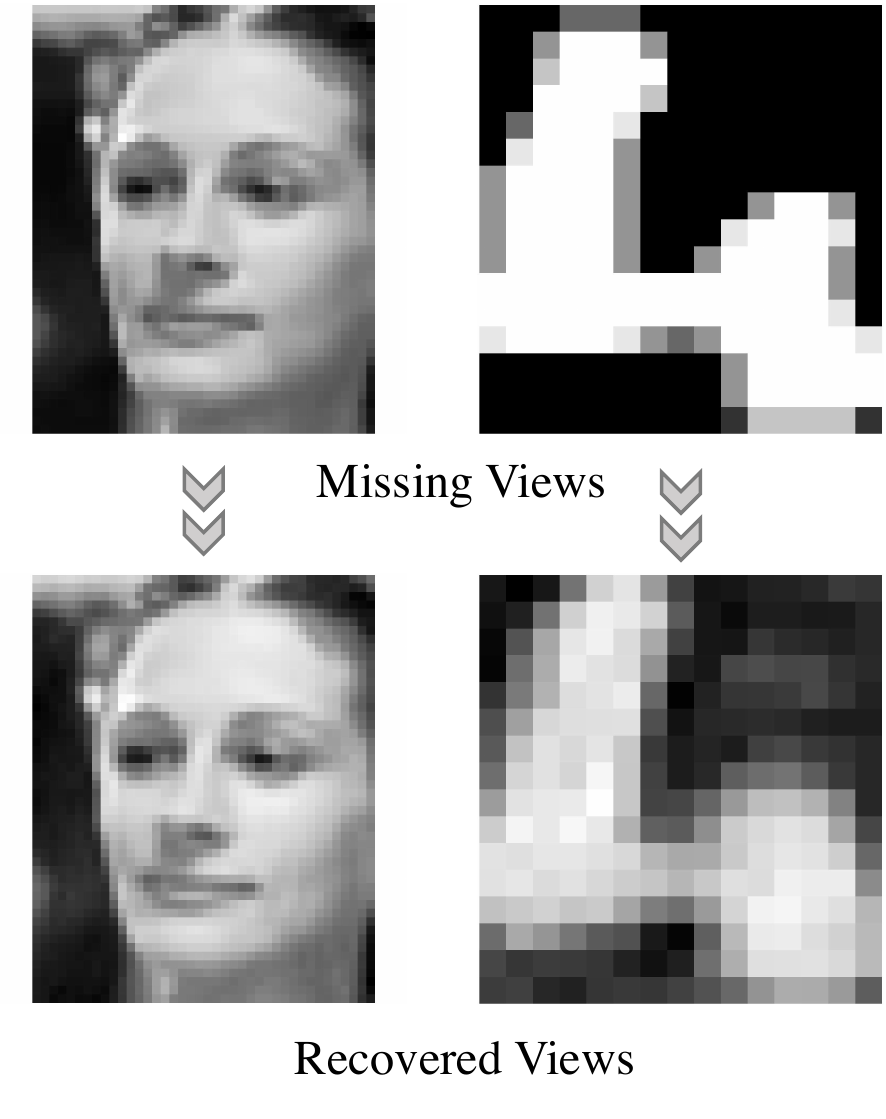}
		\caption{Missing view completion.}
	\end{center}
	\vspace{-0.5cm}
\end{figure}
In the area of incomplete multi-view learning, there are two main technical approaches to missing multi-view data, \textit{i.e.}, \textit{detour}--skip unavailable views by prior missing information or \textit{completion}--recover missing data according to existing information \cite{chen2022augmented}. In general, it is more difficult to get good recovery quality because the information content is indeed incomplete, so most methods choose to skip these missing data to avoid a bad impact. Classical partial multi-view clustering (PMVC) \cite{li2014partial} is one of the most representative IMC methods, which tries to connect different view spaces employing the samples with all views. On this basis, IMG (incomplete multi-modality grouping) \cite{zhao2016incomplete} constructs a complete Laplacian graph with the common representation in the latent space to provide the global property to subspace learning. However, both of these two grouping methods require at least one complete sample in the multi-view database. Online multi-view clustering (OMVC) \cite{shao2016online} and One-pass IMC (OPIMC) \cite{hu2019one} introduce a weight matrix with missing prior information to perform multi-view weighted matrix factorization, assigning low weights to missing views to reduce the negative impact caused by mean padding or zero padding. Liu et al. proposed a self-representation subspace clustering algorithm in incomplete multi-view setting (IMSR), which is able to learn a consistent self-representation matrix across views for clustering \cite{liu2021self}.

Although these methods avoid the effects of missing views to some extent, however, such detour or neglect is only the second-best solution to the incomplete issue \cite{liu2023self}. Some researchers attempt to recover missing information in various ways to perform complete multi-view clustering. Rai et al. developed the kernel canonical correlation analysis with incomplete views (KCCA-IV) to restore the kernel matrix of incomplete view based on the complete kernel, which requires at least one view to be intact \cite{rai2010multiview}. The multiple kernel k-means with incomplete kernels (MKKM-IK-MKC) \cite{liu2019multiple} fuses kernel completion and clustering into one framework. Wen et al. proposed an IMC framework named unified embedding alignment framework (UEAF), which emphasizes the importance of recovering missing views, and introduces both forward and reverse graph regularization to facilitate view recovery and cross-view consensus representation learning \cite{wen2019unified}.

Considering the above issues, in this paper, we aim to propose a general information recovery-driven IMC framework, which can handle arbitrary view-missing situations and enhance the performance of representation learning through efficient view completion. Meanwhile, inspired by the successful application of deep learning, especially the transformer \cite{vaswani2017attention}, we combine the characteristics of the multi-head self-attention mechanism and multi-view learning to design a transformer-style cross-view autoencoder network. Compared with a simple linear encoder, it can extract high-level semantic features and support cross-view information interaction, which is conducive to mining the complementarity of multiple views. At the same time, we skillfully integrate multi-view fusion representation learning and missing view recovery into a unified framework. More importantly, we propose a structure-aware module (recurrent graph constraint) to push the reconstructed data inversely to participate in the representation learning process, where they are allowed to collaborate with each other to achieve better clustering. Finally, we named our framework \textbf{RecFormer} and our contributions can be summarized as follows:
\begin{itemize}
	\item We design a general IMC framework named RecFormer, which includes an encoder module that integrates cross-view information interaction and high-level semantic feature extraction, and a multi-view reconstruction and recovery module based on controlled coding.
	
	\item  We propose an innovative recurrent graph embedding constraint based on an approximately complete graph generated from imputation data, which cyclically facilitates reliable feature extraction and view recovery.
	
	\item A cluster-friendly two-stage training strategy is presented in detail, and extensive experiments and intuitive visualization results demonstrate the effectiveness of our RecFormer.
\end{itemize}
\section{Preliminary}
\subsection{Problem definition and notations}
For ease of expression, we first give a formal definition of the investigated problem. Given the multi-view data $\{X^{(v)}\in \mathbb{R}^{n\times d_v}\}_{v=1}^{m}$ with $m$ views and $n$ samples, our goal is to divide them into $c$ cluster centroids. $d_v$ is the dimension of view $v$ and the dimension of embedding feature is $d_e$. In our method, a missing indicator matrix is introduced, \textit{i.e.}, $W$, whose element $W_{i,j}=1$ denotes the $j$-th view of $i$-th sample is available, otherwise  $W_{i,j}=0$. $\bar{X}^{(v)} \in \mathbb{R}^{n\times d_v}$ represents the reconstructed data of $v$-th view including missing instances and $X'^{(v)}\in \mathbb{R}^{n\times d_v}$ is the imputation view $v$ filled with recovered data. $\bm{\mathsf{Z}}\in \mathbb{R}^{n\times m\times d_e}$ is the extracted embedding tensor and its fusion representation $\bar{Z}\in \mathbb{R}^{n\times d_e}$ is our objective matrix. Following \cite{goodfellow2016deep}, all subscript representations of matrices or tensors conform to the recommended criteria. For example, $X^{(v)}_{i,:}$ denotes $i$-th row of matrix $X^{(v)}$ (\textit{i.e.}, the feature of $i$-th sample) and $\bm{\mathsf{Z}}_{:,j,:}$ represents $j$-th element of the second dimension in tensor $\bm{\mathsf{Z}}$ (\textit{i.e.}, $j$-th view of all samples). We show the main notations in Table \ref{tab:not}.
\begin{table}[!t]
\caption{Main notations used in this paper.}
\label{tab:not}%
\centering
\small

\resizebox{\linewidth}{\height}{
	\begin{tabular}{c|c}
	    \toprule
	    $n$ & \makecell[c]{the number of samples} \\
	    \midrule
	    $c$&\makecell[c]{the number of classes}\\ 
   		\midrule
   		$m$ & \makecell[c]{the number of views} \\
   		\midrule
   		$d_v$&\makecell[c]{the dimension of view $v$}\\ 
		\midrule
		$d_e$&\makecell[c]{the dimension of embedding feature}\\ 
		\midrule
	    $X^{(v)}$ & \makecell[c]{original data of view $v$ ($\mathbb{R}^{n\times d_v}$)} \\
	    \midrule
	    ${Y}$&\makecell[c]{label matrix ($\{0,1\}^{n\times c}$)}\\ 
		\midrule
	    $\bar{X}^{(v)}$ & \makecell[c]{reconstructed data of view $v$ ($\mathbb{R}^{n\times d_v}$)}\\
	    \midrule
	    ${X}'^{(v)}$ & \makecell[c]{approximately complete data of view $v$ ($\mathbb{R}^{n\times d_v}$)}\\
	    \midrule
	    $W$&  \makecell[c]{missing-view indicator ($\{0,1\}^{n\times m}$)}\\ 
	    \midrule
	    $\bm{\mathsf{Z}}$ & \makecell[c]{embedding tensor ($\mathbb{R}^{n\times m\times d_e}$)}\\
	    \midrule
	    $\bar{Z}$ & \makecell[c]{fusion feature of all views ($\mathbb{R}^{n\times d_e}$)}\\
	    \midrule
		$G^{(v)}$ & \makecell[c]{approximately complete graph ($\mathbb{R}^{n\times n}$)}\\
		\midrule
		$\text{diag}(\cdot)$ & \makecell[c]{the operator that changing a vector to diagonal matrix}\\
	\bottomrule
	\end{tabular}}
\end{table}
\subsection{Related work: UEAF}
In this subsection, we simply introduce a related IMC method: UEAF \cite{wen2019unified}. Like most IMC methods, UEAF aims to learn a consensus representation ($P$ in this work) for all views. To do this, Wen et al. developed a complex framework that integrates view recovery, consensus learning, and two graph constraint terms into one model: 
\begin{equation}
	\label{eq.ueaf}
	\begin{aligned}
	&\min_{\Upsilon}\sum_{v=1}^{m}(\alpha_{v})^{r}\left(
	\begin{aligned}
		&\|X^{(v)}+E^{(v)}Z^{(v)}-U^{(v)}P^{(v)}\|_F^2\\
		&+\lambda_1\text{trac}({E^{(v)T}L^{(v)}E^{(v)}})\\
		&+\frac{\lambda_2}{2}\sum_{i,j}^{n}\|U^{(v)}P_{:,i}-U^{(v)}P_{:,i}\|_2^2S^{2}_{i,j}
	\end{aligned}
	\right),\\
	&\begin{aligned}
		s.t. &U^{(v)T}U^{(v)} =I, \sum_{v=1}^{m}\alpha_{v}=1,\alpha_{v}\ge 0, \forall{i}, S_{i,:}\textit{1}=\textit{1},\\
		&0\leq S_{i,j}\leq 1, S_{i,i}=0, rank(L_s)=n-c.
	\end{aligned}
	\end{aligned}
\end{equation}
Such a complex model contains many variables, so we only explain the core parts in detail. $E^{(v)}$ is the error matrix used to model the missing instances and $Z^{(v)}$ is the transition matrix from missing instances to complete views. $U^{(v)}$ and $P^{(v)}$ are the basis matrix and low-dimensional representation of view $v$. $\alpha_v$ is the weight of $v$-th view in the population and $r$ is the smooth parameter. $\text{trac}(\cdot)$ denotes the trace operator. The most commendable part is the two Laplacian constraint terms, \textit{i.e.}, the Laplacian matrix $L^{(v)}$ constructed by the nearest neighbor graph is introduced to constrain the inference of missing views and another nearest neighbor graph $S$ is utilized reversely to align all recovered incomplete views. $\Upsilon$ is the optimization objective concerning the above variables.
\begin{figure*}[t!]
	\begin{center}
		\includegraphics[width=0.99\linewidth]{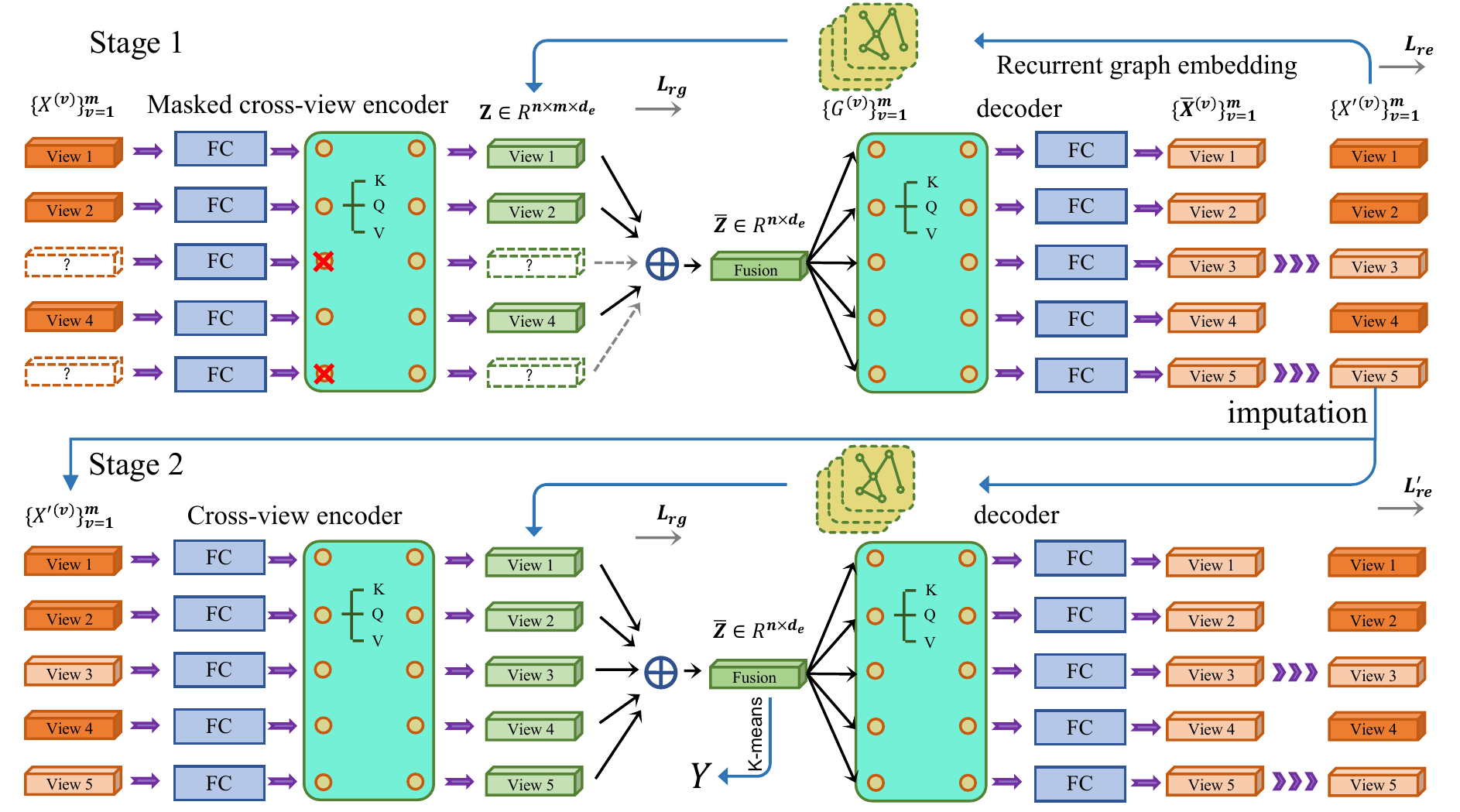}
		
	\end{center}
	\caption{Main framework of our model. The FC module means Fully Connected layer and there is a mask operation in the first cross-view encoder. In Stage 1, incomplete multi-view data is input to the encoder to extract view-specific embedding features $\bm{\mathsf{Z}}$, and these incomplete features are fused to a complete fused representation $\bar{Z}$. Then, the decoder reconstructs original features to obtain the predicted missing views. In stage 2, the entire model performs complete multi-view clustering to obtain the final result $Y$.}
	\label{fig.main}
\end{figure*}
\section{Method}
\subsection{Motivation}
As mentioned in the introduction, recovering the missing data must be based on existing information. We consider this question from the key property of multi-view data. As we know, different views enjoy same high-level semantic information in the clustering task, \textit{i.e.}, they're different descriptions of the same abstract target. If we can capture the shared high-level semantic information, then it becomes possible to infer the missing information backwards based on the learned patterns. From another perspective, missing data inference can be regarded as a generation task, which is usually implemented via autoencoder networks. Inspired by above analysis, we design a cross-view autoencoder as our main framework, whose encoder learns the high-level semantic representations and decoder attempts to recover the missing views from a fusion representation. 

Another thing that should not be ignored is that the intrinsic structure of the data is crucial for unsupervised learning, which has been demonstrated by numerous studies \cite{zang2023self}. The classical nearest neighbor graph constraint, widely used in various traditional machine learning methods, enables the extracted semantic representation to maintain the original topology of data, which not only facilitates the learning of clustering structure to a large extent but also drives the model to ‘guess’ the missing data in a more reasonable direction. Nevertheless, it should be noted that it is hard for us to directly obtain a complete graph from the incomplete data unless we can provide approximately intact data. So far, re-examining the above motivations, our idea to missing data imputation is gradually blossomed, that is, fusing approximate graph construction and missing view recovery in an unified autoencoder framework. 

At the same time, it must be considered that our downstream task is unsupervised multi-view clustering, which means that view imputation serves the complete multi-view clustering. To do this, a 2-Stage training strategy is applied in our network:
\begin{itemize}
\item \textit{Stage 1}: Missing view recovery.
\item \textit{Stage 2}: Learning multi-view clustering representation based on recovered data. 
\end{itemize}
In Stage 1, the noisy data in missing positions must be addressed to avoid the negative influence, but in Stage 2, the new imputed data (approximately complete data), consisting of the recovered views in Stage 1 and raw incomplete data, is used as the input, \textit{i.e.}, $X'^{(v)} = \text{diag}(W_{:,v})X^{(v)}+\text{diag}(1-W_{:,v})\bar{X}^{(v)}$, so the entire model can be treated as a common multi-view clustering network without any extra measures in Stage 2. It is worth noting that the input to Stage 2, $X'^{(v)}$, is fixed after training Stage 1 because frequently changing the input data makes it difficult for the model to converge. The remainder of this section elaborates on the details of our model and the main framework is shown in Figure \ref{fig.main}.

\subsection{Cross-view encoder}
It is well known that the feature dimensions of raw data are diverse, so it is difficult for a model to fuse multi-view features in the original feature space. Traditional approaches usually utilize the autoencoders, mainly composed of Fully Connected (FC) layers, to extract view-specific features, which not only aligns the different dimensions in a common space, but also maps high-dimensional features to a relatively low-dimensional space to facilitate subsequent representation learning. However, such high-level features only extracted from their corresponding raw views lack information interaction among views. Specifically, different views describe the objects from different perspectives, so each view can be considered both unique and biased. These differences (\textit{i.e.}, complementary information) naturally make multi-view data more expressive than single-view data, and how to make full use of complementary information is always one of the key problems in the field of multi-view learning. To do this, we design a transformer-style cross-view encoder with multi-view information aggregation. At first, we need a group of low-level feature extractors to project various views into a common feature space, which allows subsequent modules to process the representations of all views in parallel. For simplicity, we select $m$ Fully Connected layers as the low-level feature extractors: $\big\{\Phi_v(X^{(v)} ; \theta_{v})=\widehat{X}^{(v)}\in \mathbb{R}^{n\times d_e}\big\}_{v=1}^{m}$, where $\theta_{v}$ denotes the parameters of extractor $\Phi_v$. Notably, $\widehat{\bm{\mathsf{X}}}=\big\{\widehat{X}^{(v)}\in \mathbb{R}^{n\times d_e}\big\}_{v=1}^{m}\in \mathbb{R}^{n\times m\times d_e}$ is the input tensor of the subsequent self-attention module with no information interaction. And then, the transformer-style view aggregation module is defined as follows: Given the number of heads $h$ and the embedding tensor $\widehat{\bm{\mathsf{X}}}_{i,:,:}$ of sample $i$, we utilize $h$ groups of linear layers with weights $\big\{W^{q_t}, W^{k_t}, W^{v_t}\big\}_{t=1}^{h}$ to obtain corresponding query, key, and value maps $\{Q^t, K^t, V^t\}_{t=1}^{h}\in \mathbb{R}^{m\times d_h}$. An entire embedding feature is split into $h$ segments with dimension $d_h=d_e/h$. For head $t$ of sample $i$, we compute their score matrix as follows: 
\begin{equation}
	S^{t} = \textit{softmax}{(Q^t(K^t)^T/\sqrt{d_h})}.
\end{equation}
We need to highlight that this $S^{t}\in \mathbb{R}^{m\times m}$ is for the complete views used in Stage 2. As for the incomplete views in missing view inference stage (\textit{i.e.} Stage 1), we define the masked attention scores $\widehat{S}^{t}\in \mathbb{R}^{m\times m}$ by:
\begin{equation}
	\widehat{S}^{t }=\textit{softmax}(\textit{zerofill}(Q^t(K^t)^T/\sqrt{d_h}, W_{i,:}^TW_{i,:})),
\end{equation}
where the \textit{zerofill}$(matrix1, matrix2)$ represents the operation to replace $matrix1$'s value with -1e9 according to the position of element that equals to 0 in $matrix2$. This aims to ignore the missing views in the computation of cross-view attentions. Then, we aggregate all views by:
\begin{equation}
	A^{t} = \Bigg\{
	\begin{aligned}
		\widehat{S}^{t}V^t, \text{Stage 1}\\
		S^{t}V^t, \text{Stage 2}
	\end{aligned},
\end{equation}
where $A^t\in \mathbb{R}^{m\times d_h}$ is the new embedding features with information interaction of head $t$. Similar to the multi-head transformer, the final embedding features of $m$ views can be calculated as $A = Concat(A^1,A^2,...,A^h)\in \mathbb{R}^{m\times d_e}$. Besides, we sequentially input the $A$ into a linear layer, layer norm module, and a multilayer perceptrons to get final embedding features $Z\in \mathbb{R}^{m\times d_e}$ for each sample ($\bm{\mathsf{Z}}\in \mathbb{R}^{n\times m\times d_e}$ for all samples), \textit{i.e.}, cross-view encoder $\mathcal{E}:\{X^{(v)}\}_{v=1}^{m}\rightarrow \bm{\mathsf{Z}}$.

\subsection{View recovery based on controlled encoding}
As mentioned in the previous subsection, all views exchange information in the extraction of high-level semantic representations. However, no appropriate constraint be imposed in the extraction of encoder, which means that the encoding process is uncontrolled. To solve this problem, we exploit a simple but effective approach, named multi-view weighted fusion, to obtain a common representation of all views, which is expected to comprehensively characterize the sample across views. Specifically, our multi-view weighted fusion is formulated as follows:
\begin{equation}
	\label{eq.fusion}
	\bar{Z}_{i,:}=\sum^m_{v=1}\frac{\bm{\mathsf{Z}}_{i,v,:}W_{i,v}}{\sum_vW_{i,v}},
\end{equation}
where vector $\bar{Z}_{i,:}$ denotes the $i$-th sample of the fusion matrix $\bar{Z}\in \mathbb{R}^{n\times d_e}$. Obviously, $\bar{Z}$ in Eq. (\ref{eq.fusion}) is designed for Stage 1, and $\bar{Z}$ can be simply calculated by $\bar{Z}_{i,:}=\frac{1}{m}\sum_v\bm{\mathsf{Z}}_{i,v,:}$ in Stage 2. Furthermore, the common representation $\bar{Z}$ is required to contain all information of multiple views. To the end, a symmetrical cross-view decoder module $\mathcal{D}$ is concatenated to $\bar{Z}$ to reconstruct all the data including missing views, \textit{i.e.}, $\mathcal{D}: \{\bar{Z}\}^m\in \mathbb{R}^{n\times m\times d_e} \rightarrow \{{\bar{X}^{(v)}}\in \mathbb{R}^{n\times d_v}\}_{v=1}^m$. But in fact, due to the lack of supervisory information to directly discriminate the recovered data, we can only leverage the available original data to impose a partial reconstruction constraint. In other words, this recovery is a natural but necessary byproduct of the proposed autoencoder framework that aims to learn the common representation shared by the available views. As a result, we introduce a weighted reconstruction loss $\mathcal{L}_{re}$:
\begin{equation}
	\mathcal{L}_{re}= \frac{1}{mn}\sum_{v=1}^m\sum_{i=1}^n\frac{1}{d_v}\big\|\bar{X}_{i,:}^{(v)}-X_{i,:}^{(v)}\big\|_2^2W_{i,v},
	\label{eq.lossre1}
\end{equation}
where $\bar{X}_{i,:}^{(v)}$ is the $i$-th sample of the reconstructed data $\bar{X}$. And in Stage 2, the $\mathcal{L}_{re}$ degenerates into 
\begin{equation}
	\mathcal{L}_{re}'= \frac{1}{mn}\sum_{v=1}^m\sum_{i=1}^n\frac{1}{d_v}\big\|\bar{X}_{i,:}^{(v)}-X_{i,:}'^{(v)}\big\|_2^2.
	\label{eq.lossre2}
\end{equation}

\subsection{Recurrent graph constraint}
In recent years, researchers have been accustomed to adding graph constraints to traditional multi-view learning methods, which help preserve the original intrinsic structure of data by constructing a prior adjacency matrix. This is based on this basic manifold assumption: if two samples are close to each other in original feature space, then they are also close in the embedding space \cite{wen2020adaptive}. But in the case of incomplete data, some existing methods simply skip the missing views to construct the adjacency graph, which is obviously biased, especially on the databases with lager missing rates. Therefore, we expect to obtain an approximately complete adjacency graph to guide the encoder for the extraction of high-level semantic features. On the other hand, more discriminative semantic features can also facilitate the recovery of missing views and thus help to construct a more realistic graph. Combining these two points, we innovatively propose the recurrent graph constraint:
\begin{equation}
	\label{eq.graph1}
	\mathcal{L}_{rg} = \frac{1}{mn^2}\sum_{v=1}^{m}\sum_{i=1}^{n}\sum_{j=1}^{n}\big\|\bm{\mathsf{Z}}_{i,v,:}|_k-\bm{\mathsf{Z}}_{j,v,:}|_k\big\|_2^2G^{(v)}_{i,j}|_{k-1},
\end{equation}
where $\big\{G^{(v)}\in \mathbb{R}^{n\times n}\big\}_{v=1}^{m}$ are the approximately complete graphs generated from imputation data of last epoch, \textit{i.e.}, $G^{(v)}|_{k-1}=\textit{knn}(X'^{(v)}|_{k-1},K)$ and ‘$|_k$’ denotes the $k$-th epoch. $K$ is the number of nearest neighbors. $G^{(v)}_{i,j}=1$ means instance $j$ is one of the $K$-neighbors of instance $i$ in view $v$. The $\mathcal{L}_{rg}$ is executed only if $k>1$. In fact, deep learning models are usually trained in mini-batch iterations to reduce memory overhead, and ours is no exception. But in doing so, the neighbor graph contains only a small batch of samples, which means that the graph constraint is local rather than global. In order to balance the mini-batch training approach and the intact graph constraint, we rewrite Eq. (\ref{eq.graph1}) in the case of mini-batch training as follows: 
\begin{equation}
	\label{eq.graphloss}
	\mathcal{L}_{rg}^{batch} = \frac{1}{mnb}\sum_{v=1}^{m}\sum_{i=1}^{b}\sum_{j=1}^{n}\big\|\bm{\mathsf{Z}}^{batch}_{i,v,:}|_k-\bm{\mathsf{Z}}_{j,v,:}|_{k-1}\big\|_2^2G^{(v)}_{i,j}|_{k-1},
\end{equation}
where the size of mini-batch is $b$, and $\bm{\mathsf{Z}}^{batch}\in \mathbb{R}^{b\times m \times d_e}$ is the output of the cross-view encoder corresponding to each mini-batch data. In the computation of graph loss, to maximize the use of global structural information, we preserve all the embedding tensor $\bm{\mathsf{Z}}$ obtained in the last epoch and update its corresponding part after the current batch is processed. Notably, in Stage 2, the graph $G$ from Stage 1 is fixed rather than updated by the last output, \textit{i.e.}, $G^{(v)}|_{k-1}=G^{(v)}|_{k}$.

\subsection{Overall loss function and algorithm}
To sum up, our overall loss function in Stage 1 is:
\begin{equation}
	\label{eq.loss1}
	\mathcal{L} = \mathcal{L}_{re}+\beta\mathcal{L}_{rg},
\end{equation}
$\beta$ is penalty parameter to balance the two losses. And the loss function in Stage 2 is:
\begin{equation}
	\label{eq.loss2}
	\mathcal{L}' = \mathcal{L}'_{re}+\beta\mathcal{L}_{rg}.
\end{equation}
As mentioned above, we conduct complete multi-view clustering in Stage 2, and the fusion embedding feature $\bar{Z}\in \mathbb{R}^{n\times d_e}$ obtained in Stage 2 is regarded as our clustering indicator matrix. For simplicity, we perform \textit{K}-means \cite{macqueen1967classification} on the $\bar{Z}$ to obtain final clustering results $P \in \mathbb{R}^{n\times c}$. Training process of RecFormer is shown in Algorithm \ref{al.1}. It is important to state that the recovered data is only utilized to update the approximately complete graph for the following epoch instead of being fed into the network as the input feature in Stage 1. And the recovered data, output by the final epoch of Stage 1, is used as the input of model in the entire Stage 2. Because we argue that updating the input of the network at each epoch invalidates the network parameters and gradients trained on the previous data.

\begin{algorithm}[t]
	\caption{Training process of RecFormer}
	\label{al.1}
	\textbf{Input}: Incomplete multi-view data $\big\{X^{(v)}\big\}_{v=1}^{m}$ with view-presence information, hyperparameters $\beta$ and $K$.\\
	\textbf{Initialization}: Construct $\left\{W^{(v)}\right\}_{v=1}^{m}$ according to the view-presence information. Initialize parameters of model. Set training epochs $e_1$ and $e_2$ for two stages.
	\begin{algorithmic}[1]
		\State $k=1$
		\While{$k\le e_1$}                           \Comment{Stage 1}                     	
		\State Compute multi-view embedding tensor $\bm{\mathsf{Z}}$.
		\If{$k> 1$}
		\State Compute graph loss $\mathcal{L}_{rg}$ by Eq. (\ref{eq.graph1}).
		\EndIf
		\State Compute fusion representation $\bar{Z}$ by Eq. (\ref{eq.fusion}).
		\State Compute reconstructed data $\big\{\bar{X}^{(v)}\big\}_{v=1}^{m}$ by decoder and reconstruction loss $\mathcal{L}_{re}$ by Eq. (\ref{eq.lossre1}).
		\State Complete missing data in $\left\{X^{(v)}\right\}_{v=1}^{m}$ with $\big\{\bar{X}^{(v)}\big\}_{v=1}^{m}$ to get new data $\big\{X'^{(v)}\big\}_{v=1}^{m}$.
		\State Generate graph $\{G^{(v)}\}_{v=1}^{m}$ using $\big\{X'^{(v)}\big\}_{v=1}^{m}$.
		\State Compute total loss $\mathcal{L}$ by Eq. (\ref{eq.loss1}) and update network parameters.
		\State $k=k+1$.
		\EndWhile
		\State Update $\big\{X^{(v)}\big\}_{v=1}^{m}$ with $\big\{X'^{(v)}\big\}_{v=1}^{m}$ and fix $\{G^{(v)}\}_{v=1}^{m}$.
		\While{$k\le e_2$}                           \Comment{Stage 2}                     	
		\State Compute multi-view embedding tensor $\bm{\mathsf{Z}}$.
		\State Compute graph loss $\mathcal{L}_{rg}$ by Eq. (\ref{eq.graph1}).
		\State Compute fusion representation $\bar{Z}$ by Eq. (\ref{eq.fusion}).
		\State Compute reconstructed data $\big\{\bar{X}^{(v)}\big\}_{v=1}^{m}$ by decoder and reconstruction loss $\mathcal{L}'_{re}$ by Eq. (\ref{eq.lossre2}).
		\State Compute total loss $\mathcal{L}'$ by Eq. (\ref{eq.loss2}) and update network parameters.
		\State $k=k+1$.
		\EndWhile
		\State Run \textit{K}-means on $\bar{Z}$ to obtain final prediction $P$.
	\end{algorithmic}
	\textbf{Output}: $P$
\end{algorithm}

\section{Experiments}
\subsection{Experimental settings}
\textbf{Databases}: In order to evaluate the performance of our model, we adopt five popular databases in our comparison experiments: 

\begin{itemize}
\item \textit{\textbf{Handwritten digit}} \cite{asuncion2007uci}, a dataset widely used in various fields, contains 2000 handwritten digital images with ten classes from '0' to '9'. There are five categories of features as 5 views selected in our experiments. The first view represents a linear combination of pixels from the original picture with the size of $16\times 15$. 
\item \textit{\textbf{Caltech7}} \cite{cai2013multi} is a subset of the Caltech101 database \cite{fei2004learning}, and we select 1474 images covering seven categories from it. Each image is extracted with 6 types of features, i.e, LBP, Gist, Hog, Cenhist, Gabor, and wavelet-moments. 
\item \textit{\textbf{NH\_face}} \cite{cao2015constrained}, as a subset of the NH database, is composed of 4660 images belonging to five persons in the movie 'Notting Hill' \cite{wu2013constrained}. 3 views in terms of Gabor, gray pixels (size of $40\times 50$), and LBP are included in the NH\_face database. 
\item \textbf{\textit{Animal}} \cite{fei2004learning,zhang2019cpm} is a larger dataset with up to 10158 images and 50 categories, whose features are extracted by DECAF \cite{krizhevsky2017imagenet} and Vgg19 \cite{simonyan2014very}. All samples and features are adopted in our experiments. 
\item \textbf{\textit{Aloi\_deep}}\footnote{Aloi\_deep database: \url{https://github.com/justsmart/Recformer}} is a new multi-view database proposed by this paper, which is derived from the Aloi database \cite{geusebroek2005amsterdam}. The original Aloi database contains 110250 images of 1000 small objects. We select 100 objects as 100 categories, each with about 108 images, for a total of 10800 images. And we utilize three typical deep neural networks, \textit{i.e.}, ResNet50 \cite{he2016deep}, Vgg16 \cite{simonyan2014very}, and Inception-v3 \cite{szegedy2016rethinking} with pre-trained weights, to extract the three-view features. 
\end{itemize}
More information about the five datasets is listed in Table \ref{tab:dataset}.

\begin{table}[tbp!]
	\caption{Detailed information about five multi-view databases.}
	\label{tab:dataset}
	
	\centering
	\begin{tabularx}{\linewidth}{ccccc}
		\toprule
		Database   & \# Class & \# View & \# Samples &      \# Features       \\ \midrule\midrule
		Handwritten &    10    &    5    &    2000    &    76/216/64/240/47    \\
		Caltech7   &    7     &    6    &    1474    & 48/40/254/1984/512/928 \\
		NH\_face   &    5     &    3    &    4660    &     6750/2000/3304     \\
		Animal    &    50    &    2    &   10158    &       4096/4096        \\ 
		Aloi\_deep    &    100   &    3    &   10800    &       2048/4096/2048        \\ \bottomrule
	\end{tabularx}

\end{table}
\textbf{Preprocessing of incomplete datasets}: To generate the incomplete datasets to simulate the missing-view case, following \cite{liu2022localized,wen2022survey}, we randomly disable [10\%, 30\%, 50\%, 70\%] of the instances of each view but keep at least one view available for each sample. As for the Animal dataset with only two views, we randomly select [10\%, 30\%, 50\%] of all samples as the paired samples with two views. The first view is removed for half of the remaining samples, and the second view is removed for the other half. In the experiments, all removed data is populated with '0' to hold its full shape.

\textbf{Comparison methods}: In our experiments, eleven state-of-the-art methods are selected to evaluate the performance of the proposed RecFormer, of which \textbf{OMVC} \cite{shao2016online}, \textbf{OPIMC} \cite{hu2019one}, \textbf{MKKM-IK-MKC} \cite{liu2019multiple}, \textbf{UEAF} \cite{wen2019unified}, and \textbf{IMSR} \cite{liu2021self} have been described in the introduction. The other six comparison methods are as follows: 
\begin{enumerate}
\item \textbf{BSV} \cite{zhao2016incomplete}, a simple baseline method, fills missing views with the average vector and performs \textit{K}-means on each view to obtain the best result.
\item \textbf{Concat} \cite{zhao2016incomplete} is another popular baseline method, which aligns all views with the same imputation strategy as BSV and simply concatenates them to conduct single-view clustering.
\item \textbf{MIC} \cite{shao2015multiple}, based on non-negative matrix factorization technology, also introduces the prior missing information to help learn robust latent representations.
\item \textbf{DAIMC} \cite{hu2019doubly} not only introduces a prior matrix to avoid the negative impact of missing instances, but also additionally introduces a regression coefficient matrix to align the basis matrix of individual views in the latent space. 
\item \textbf{OS-LF-IMVC} \cite{zhang2021one} integrate the imputation of missing views and final clustering into one optimization procedure. It outputs the clustering results directly without using k-means.
\item \textbf{HCP-IMSC} \cite{li2022high} aims to preserve the high-order geometric structure of the incomplete multi-view data by constructing a hypergraph from the self-representation matrix.

\end{enumerate}
\begin{table*}[!t]
	
	\centering
	\caption{Results on \textbf{Handwritten} database with different incomplete rates. The 1st and 	2nd best results are marked in \textbf{bold} and \underline{underline}.}

	\resizebox{0.99\textwidth}{!}{
		\begin{tabular}{c|ccc|ccc|ccc}
			\toprule
			
			\multirow{2}{*} {Method} & \multicolumn{3}{c|}{\textbf{ACC (\%)}}      & \multicolumn{3}{c|}{\textbf{NMI (\%)}}      & \multicolumn{3}{c}{\textbf{Purity (\%)}} \\
			\cmidrule{2-10}
			~& 30\% & 50\% & 70\% &30\% & 50\% & 70\% &30\% & 50\% & 70\% \\
			\midrule
			BSV	&51.49$\pm$2.29	&38.24$\pm$2.25	&27.15$\pm$1.31	&47.01$\pm$1.71	&32.21$\pm$1.00	&19.48$\pm$0.69	&53.69$\pm$1.54	&39.54$\pm$2.04	&27.76$\pm$1.09\\
			Concat	&55.48$\pm$1.57	&42.19$\pm$0.99	&28.31$\pm$0.75	&51.66$\pm$0.99	&38.24$\pm$1.59	&23.50$\pm$0.95	&57.32$\pm$1.15	&44.21$\pm$0.98	&30.45$\pm$0.80\\
			MIC	&73.29$\pm$3.41	&61.27$\pm$3.16	&41.34$\pm$2.69	&65.39$\pm$2.08	&52.95$\pm$1.33	&34.71$\pm$2.11	&74.31$\pm$3.15	&62.89$\pm$3.08	&43.25$\pm$2.86\\
			DAIMC	&86.73$\pm$0.79	&\underline{81.92$\pm$0.88}	&60.44$\pm$6.87	&76.65$\pm$1.07	&68.77$\pm$0.99	&47.10$\pm$4.79	&86.73$\pm$0.79	&\underline{81.92$\pm$0.88}	&61.24$\pm$0.42\\
			OMVC	&55.00$\pm$5.06	&36.40$\pm$4.93	&29.80$\pm$4.63	&44.99$\pm$4.56	&35.16$\pm$4.62	&25.83$\pm$8.37	&55.89$\pm$4.72	&38.51$\pm$4.87	&31.95$\pm$5.22\\
			OPIMC	&76.45$\pm$5.15	&69.50$\pm$6.54	&56.66$\pm$10.06	&73.74$\pm$3.42	&66.57$\pm$4.18	&51.86$\pm$7.97	&78.96$\pm$3.37	&72.00$\pm$6.39	&58.16$\pm$10.35\\
			MKKM-IK-MKC	&69.07$\pm$0.73	&66.08$\pm$3.25	&55.55$\pm$1.39	&65.42$\pm$0.61	&59.04$\pm$2.69	&47.36$\pm$1.78 &73.12$\pm$0.61	&66.58$\pm$3.26	&56.26$\pm$1.07\\
			UEAF	&76.11$\pm$7.74	&65.39$\pm$5.09	&61.11$\pm$1.41	&69.37$\pm$3.31	&55.09$\pm$2.05	&50.56$\pm$1.11	&76.51$\pm$7.17	&66.49$\pm$4.18	&61.60$\pm$1.09\\
			IMSR & \underline{89.74$\pm$1.02} & 83.68$\pm$1.66 & 62.10$\pm$4.59 & \underline{81.57$\pm$1.14} & \underline{72.81$\pm$2.04} & 53.92$\pm$2.86 & \underline{89.74$\pm$1.02} & 83.68$\pm$1.66 & 62.42$\pm$4.71 \\ 
   			OS-LF-IMVC  & 70.21$\pm$2.10 & 55.17$\pm$3.09 & 35.79$\pm$2.82  & 60.98$\pm$1.21 & 44.70$\pm$0.86 & 27.07$\pm$1.46  & 70.87$\pm$1.58 & 56.76$\pm$2.39 & 37.17$\pm$2.71  \\ 
       		HCP-IMSC & 78.33$\pm$0.97 & 76.19$\pm$1.53 & \underline{66.23$\pm$2.49} & 72.84$\pm$0.83 & 68.81$\pm$0.72 & \underline{56.79$\pm$2.26} & 78.73$\pm$0.57 & 76.32$\pm$1.54 & \underline{68.24$\pm$3.26} \\ 
			Ours	&\textbf{93.07$\pm$0.41}	&\textbf{91.74$\pm$0.43}	&\textbf{84.43$\pm$1.18}	&\textbf{86.12$\pm$0.64}	&\textbf{83.39$\pm$0.91}	&\textbf{72.23$\pm$1.08}	&\textbf{93.07$\pm$0.41}	&\textbf{91.74$\pm$0.43}	&\textbf{84.43$\pm$1.18}\\
			\bottomrule
		\end{tabular}
	}
	\label{tab:res1}
\end{table*}
\begin{table*}[!t]
	
	\centering
	\caption{Results on \textbf{Caltech7} database with different incomplete rates. The 1st and 2nd best results are marked in \textbf{bold} and \underline{underline}.}

	\resizebox{0.99\textwidth}{!}{
		\begin{tabular}{c|ccc|ccc|ccc}
			\toprule
			
			\multirow{2}{*} {Method} & \multicolumn{3}{c|}{\textbf{ACC (\%)}}      & \multicolumn{3}{c|}{\textbf{NMI (\%)}}      & \multicolumn{3}{c}{\textbf{Purity (\%)}} \\
			\cmidrule{2-10}
			~& 10\% & 30\% & 50\% &10\% & 30\% & 50\% &10\% & 30\% & 50\% \\
			\midrule
			BSV	&43.89$\pm$1.37	&39.06$\pm$1.26	&38.31$\pm$1.68	&39.66$\pm$2.23	&31.63$\pm$1.51	&26.81$\pm$1.38	&84.08$\pm$1.23	&75.25$\pm$0.71	&68.97$\pm$0.49\\
			Concat	&41.25$\pm$1.67	&40.55$\pm$1.89	&38.06$\pm$0.88	&43.48$\pm$0.92	&37.99$\pm$2.17	&30.28$\pm$0.66	&\underline{84.91$\pm$0.50}	&82.54$\pm$1.12	&77.56$\pm$0.98\\
			MIC	&44.07$\pm$4.97	&38.01$\pm$2.12	&35.80$\pm$2.34	&33.71$\pm$2.66	&27.35$\pm$1.69	&20.44$\pm$0.98	&78.12$\pm$1.76	&73.31$\pm$0.72	&68.26$\pm$1.40\\
			DAIMC	&48.29$\pm$6.76	&47.46$\pm$3.42	&44.89$\pm$4.88	&\underline{44.61$\pm$3.88}	&38.45$\pm$2.88	&36.28$\pm$2.34	&83.32$\pm$1.31	&76.83$\pm$3.23	&75.50$\pm$1.17\\
			OMVC	&40.88$\pm$1.54	&36.82$\pm$1.65	&33.28$\pm$4.40	&28.13$\pm$2.54	&25.32$\pm$1.03	&18.76$\pm$4.22	&79.21$\pm$1.77	&77.73$\pm$1.35	&74.05$\pm$4.74\\
			OPIMC	&49.24$\pm$2.89	&48.34$\pm$4.36	&44.12$\pm$5.85	&42.98$\pm$1.02	&\underline{41.54$\pm$2.38} &35.98$\pm$2.77	&84.89$\pm$0.69	&\underline{83.70$\pm$1.80}	&\underline{80.64$\pm$2.06}\\
			MKKM-IK-MKC	&36.54$\pm$0.51	&34.87$\pm$1.53	&36.05$\pm$0.45	&24.09$\pm$0.98	&23.45$\pm$0.52	&22.91$\pm$0.67	&72.98$\pm$0.80	&73.82$\pm$0.53	&72.52$\pm$1.55\\
			UEAF	&50.82$\pm$4.05	&42.71$\pm$0.84	&36.32$\pm$4.22	&39.44$\pm$2.07	&31.07$\pm$1.99	&24.02$\pm$1.37	&81.49$\pm$1.78	&78.26$\pm$2.12	&76.29$\pm$1.93\\
			IMSR & \textbf{55.13$\pm$4.52} & 38.81$\pm$1.04 & 24.60$\pm$1.33 & 44.16$\pm$4.22 & 27.96$\pm$1.63 & 9.12$\pm$1.43 & 82.58$\pm$2.39 & 74.02$\pm$1.18 & 62.75$\pm$1.27 \\ 
			OS-LF-IMVC  & 42.78$\pm$0.84 & 38.93$\pm$0.26 & 34.91$\pm$0.92 & 29.26$\pm$1.98 & 24.50$\pm$1.51 & 20.78$\pm$1.52 & 77.22$\pm$1.69 & 75.36$\pm$1.11 & 74.23$\pm$2.02  \\ 
			HCP-IMSC & \underline{54.67$\pm$1.25} & \textbf{53.51$\pm$1.67} & \underline{47.38$\pm$4.23} & 41.55$\pm$0.97 & 40.54$\pm$3.01 & \underline{37.03$\pm$1.53} & 83.05$\pm$1.13 & 83.04$\pm$3.32 & 80.16$\pm$1.87 \\ 
			Ours	&53.13$\pm$2.65	&\underline{51.42$\pm$1.57}	&\textbf{51.23$\pm$3.32}	&\textbf{54.56$\pm$2.68}	&\textbf{52.48$\pm$0.83}	&\textbf{50.29$\pm$1.37}	&\textbf{85.17$\pm$0.91}	&\textbf{84.02$\pm$0.69}	&\textbf{83.79$\pm$0.45}\\
			\bottomrule
		\end{tabular}
	}
	\label{tab:res2}
\end{table*}
\begin{table*}[!t]
	
	\centering
	\caption{Results on \textbf{NH\_face} database with different incomplete rates. The 1st and 2nd best results are marked in \textbf{bold} and \underline{underline}.}

	\resizebox{0.99\textwidth}{!}{
		\begin{tabular}{c|ccc|ccc|ccc}
			\toprule
			
			\multirow{2}{*} {Method} & \multicolumn{3}{c|}{\textbf{ACC (\%)}}      & \multicolumn{3}{c|}{\textbf{NMI (\%)}}      & \multicolumn{3}{c}{\textbf{Purity (\%)}} \\
			\cmidrule{2-10}
			~& 10\% & 30\% & 50\% &10\% & 30\% & 50\% &10\% & 30\% & 50\% \\
			\midrule
			BSV	&69.09$\pm$4.76	&56.82$\pm$2.28	&46.54$\pm$1.90	&56.26$\pm$4.07	&39.29$\pm$2.63	&26.20$\pm$1.09	&73.59$\pm$2.96	&60.13$\pm$1.52	&50.15$\pm$1.28\\
			Concat	&85.87$\pm$2.64	&63.14$\pm$2.78	&52.99$\pm$1.84	&81.46$\pm$1.70	&59.12$\pm$1.14	&47.42$\pm$1.29	&87.39$\pm$1.57	& 87.39$\pm$1.57	&62.21$\pm$1.04\\
			MIC	&78.83$\pm$4.07	&77.22$\pm$0.76	&75.77$\pm$4.05	&73.04$\pm$2.78	&66.82$\pm$0.80	&62.84$\pm$3.20	&82.54$\pm$1.66
			&78.83$\pm$0.64	&77.40$\pm$3.48\\
			DAIMC	&87.42$\pm$4.15 &85.35$\pm$3.44 &84.57$\pm$3.49 &78.37$\pm$3.42 &74.71$\pm$2.91 &70.09$\pm$5.08 &87.03$\pm$2.74 &85.66$\pm$2.91 &84.66$\pm$3.41	\\
			OMVC	&75.35$\pm$2.11 &72.85$\pm$3.17 &70.61$\pm$2.77 &68.45$\pm$3.22 &65.44$\pm$2.89 &63.34$\pm$4.36 &80.89$\pm$3.05 &77.96$\pm$2.33 &74.52$\pm$3.55	\\									
			OPIMC	&79.82$\pm$8.32	&74.57$\pm$3.81	&71.25$\pm$6.27	&69.92$\pm$6.36	&66.87$\pm$1.86	&64.65$\pm$6.94	&81.56$\pm$5.12	&79.02$\pm$1.27	&78.21$\pm$4.01\\
			MKKM-IK-MKC	&74.34$\pm$0.34	&75.92$\pm$0.93	&71.22$\pm$1.19	&65.21$\pm$0.32	&66.83$\pm$1.24	&65.27$\pm$1.66	&78.96$\pm$0.07	&79.18$\pm$0.16	&79.94$\pm$1.03\\
			UEAF	&80.36$\pm$0.10 &71.22$\pm$0.68 &64.37$\pm$1.13 &67.11$\pm$0.52 &55.52$\pm$2.55 &47.97$\pm$1.50 &81.67$\pm$0.13 &73.32$\pm$0.70 &68.49$\pm$1.21	\\	
  			IMSR & \underline{91.20$\pm$1.43} & \underline{91.32$\pm$2.24} & \underline{89.93$\pm$0.80} & \underline{82.65$\pm$1.94} & \underline{81.42$\pm$3.07} & \underline{79.26$\pm$0.87} & \underline{91.20$\pm$1.43} & \underline{91.32$\pm$2.24} & \underline{89.93$\pm$0.80}  \\ 
   			OS-LF-IMVC  & 73.41$\pm$6.10 & 70.52$\pm$8.94 & 69.35$\pm$3.72 & 51.65$\pm$5.45 & 56.45$\pm$7.98 & 56.07$\pm$3.58 & 74.95$\pm$3.91 & 75.30$\pm$6.30 & 75.34$\pm$1.37  \\ 
   			HCP-IMSC & 79.20$\pm$1.13 & 79.27$\pm$0.94 & 71.25$\pm$1.41 & 73.20$\pm$1.62 & 70.63$\pm$3.32 & 61.81$\pm$2.24 & 81.26$\pm$1.27 & 81.05$\pm$1.14 & 76.48$\pm$0.94 \\ 
			Ours	&\textbf{97.10$\pm$1.02}	&\textbf{96.40$\pm$2.17}	&\textbf{95.40$\pm$4.82}	&\textbf{93.62$\pm$1.49}	&\textbf{92.03$\pm$3.80}	&\textbf{92.01$\pm$4.24}	&\textbf{97.10$\pm$1.02}	&\textbf{96.40$\pm$2.17}	&\textbf{95.48$\pm$4.66}\\
			\bottomrule
		\end{tabular}
	}
	\label{tab:res3}
\end{table*}

\begin{table*}[t!]
	
	\centering
	\caption{Results on \textbf{Animal} database with different paired rates. The 1st and 2nd best results are marked in \textbf{bold} and \underline{underline}.}

	\resizebox{0.99\textwidth}{!}{
		\begin{tabular}{c|ccc|ccc|ccc}
			\toprule
			
			\multirow{2}{*} {Method} & \multicolumn{3}{c|}{\textbf{ACC (\%)}}      & \multicolumn{3}{c|}{\textbf{NMI (\%)}}      & \multicolumn{3}{c}{\textbf{Purity (\%)}} \\
			\cmidrule{2-10}
			~& 30\% & 50\% & 70\% &30\% & 50\% & 70\% &30\% & 50\% & 70\% \\
			\midrule
			BSV	&42.05$\pm$1.20	&48.63$\pm$1.89	&56.22$\pm$1.20	&48.16$\pm$0.44	&55.91$\pm$0.58	&63.99$\pm$0.38	&45.20$\pm$0.88	&52.26$\pm$1.19	&60.31$\pm$0.78\\
			Concat	&42.79$\pm$0.67	&49.34$\pm$1.39	&53.99$\pm$0.99	&55.46$\pm$0.16	&59.31$\pm$0.38	&63.88$\pm$0.35	&48.12$\pm$0.45	&53.24$\pm$0.88	&59.26$\pm$0.81\\
			MIC	&43.38$\pm$0.63	&45.88$\pm$0.34	&49.15$\pm$0.88	&52.79$\pm$0.77	&55.69$\pm$0.36	&59.30$\pm$0.54	&49.21$\pm$0.78	&52.31$\pm$0.34	&55.33$\pm$0.64\\
			DAIMC	&50.18$\pm$2.18	&53.87$\pm$1.36	&56.42$\pm$1.37	&55.03$\pm$1.03	&59.36$\pm$1.16	&62.76$\pm$0.46	&54.82$\pm$1.57	&59.51$\pm$1.65	&62.12$\pm$1.04\\
			OMVC	&42.51$\pm$0.89	&43.98$\pm$0.77	&46.39$\pm$1.02	&50.77$\pm$0.63	&53.11$\pm$0.83	&55.38$\pm$0.46	&47.33$\pm$0.66	&50.42$\pm$0.91	&52.97$\pm$0.76\\
			OPIMC	&46.33$\pm$2.14	&53.14$\pm$1.38	&53.88$\pm$1.26	&52.34$\pm$0.69	&58.51$\pm$0.46	&62.04$\pm$0.26	&49.49$\pm$1.41	&56.23$\pm$1.20	&57.91$\pm$0.43\\
			MKKM-IK-MKC &\underline{51.77$\pm$0.48}	&\textbf{57.75$\pm$0.38}	&\textbf{61.18$\pm$0.59}	&56.54$\pm$0.33	&\underline{61.66$\pm$0.22}	&\underline{66.28$\pm$0.27} &56.14$\pm$0.48	&\textbf{62.14$\pm$0.41}	&\textbf{66.40$\pm$0.53}\\
			UEAF	&45.73$\pm$12.9	&51.86$\pm$6.48	&58.19$\pm$3.04	&51.61$\pm$12.87	&58.43$\pm$7.53	&64.92$\pm$3.95 &49.10$\pm$0.27	&55.36$\pm$0.36	&63.02$\pm$0.47\\
   			IMSR & 51.14$\pm$1.55 & 53.93$\pm$1.10 & 55.76$\pm$2.13 & \underline{56.77$\pm$0.47} & 60.43$\pm$0.96 & 63.54$\pm$1.01 & \textbf{58.08$\pm$0.90} & 61.24$\pm$1.58 & 63.90$\pm$1.35  \\ 
   			OS-LF-IMVC  & 40.53$\pm$0.58 & 48.24$\pm$1.22 & 55.07$\pm$1.44 & 50.53$\pm$0.68 & 54.49$\pm$0.25 & 58.80$\pm$0.49 & 45.31$\pm$0.92 & 51.68$\pm$1.10 & 56.93$\pm$1.22  \\ 
   			HCP-IMSC & 51.05$\pm$1.10 & 54.29$\pm$0.76 & 55.52$\pm$1.09 & 55.12$\pm$0.30 & 58.11$\pm$0.46 & 60.21$\pm$0.60 & 55.49$\pm$0.77 & 58.57$\pm$1.04 & 59.54$\pm$0.78 \\ 
			Ours	&\textbf{52.90$\pm$0.50}	&\underline{56.00$\pm$0.72}	&\underline{59.20$\pm$0.43}	&\textbf{59.65$\pm$0.30}	&\textbf{63.32$\pm$0.92}	&\textbf{66.49$\pm$0.77}	&\underline{57.78$\pm$0.11}	&\underline{61.28$\pm$0.71}	&\underline{64.12$\pm$1.07}\\
			\bottomrule
		\end{tabular}
}
	\label{tab:res4}
\end{table*}
\begin{table*}[t!]
	\centering
	\caption{Results on \textbf{Aloi\_deep} database with different incomplete rates. The 1st and 2nd best results are marked in \textbf{bold} and \underline{underline}.}

	\resizebox{0.99\textwidth}{!}{
		\begin{tabular}{c|ccc|ccc|ccc}
			\toprule
			
			\multirow{2}{*} {Method} & \multicolumn{3}{c|}{\textbf{ACC (\%)}}      & \multicolumn{3}{c|}{\textbf{NMI (\%)}}      & \multicolumn{3}{c}{\textbf{Purity (\%)}} \\
			\cmidrule{2-10}
			~& 10\% & 30\% & 50\% &10\% & 30\% & 50\% &10\% & 30\% & 50\% \\
			\midrule
			BSV & 64.14$\pm$1.23 & 50.63$\pm$2.20 & 37.37$\pm$1.34 & 81.29$\pm$0.65 & 63.15$\pm$0.62 & 45.36$\pm$0.59 & 69.89$\pm$1.13 & 55.10$\pm$1.76 & 40.18$\pm$1.07  \\ 
			Concat & 71.07$\pm$2.90 & 59.60$\pm$1.26 & 41.39$\pm$1.30 & 89.75$\pm$1.24 & 77.47$\pm$0.46 & 68.63$\pm$0.97 & 76.52$\pm$2.39 & 64.44$\pm$0.88 & 44.99$\pm$1.54 \\ 
			MIC & 43.69$\pm$2.30	&35.54$\pm$1.19	&27.96$\pm$1.39	&72.18$\pm$2.29	&66.16$\pm$2.68	&59.10$\pm$2.82	&44.77$\pm$0.31	&36.30$\pm$0.25	&28.35$\pm$0.18 \\ 
			DAIMC & 84.07$\pm$1.27 & 81.99$\pm$1.32 & 69.00$\pm$2.75 & 95.66$\pm$0.38 & 94.78$\pm$0.23 & 87.70$\pm$1.63 & 87.65$\pm$0.85 & 85.64$\pm$0.82 & 72.61$\pm$0.63  \\ 
			OMVC & 63.13$\pm$1.43 & 51.02$\pm$1.45 & 35.18$\pm$0.62 & 80.99$\pm$1.37 & 69.54$\pm$0.87 & 57.91$\pm$0.80 & 67.58$\pm$1.30 & 55.59$\pm$1.36 & 39.37$\pm$0.72  \\ 
			OPIMC & 47.09$\pm$1.77 & 35.07$\pm$1.99 & 33.97$\pm$1.64 & 77.56$\pm$1.01 & 69.05$\pm$0.79 & 67.62$\pm$1.71 & 51.17$\pm$0.37 & 36.51$\pm$0.29 & 34.73$\pm$0.28  \\ 
			MKKM-IK-MKC & 83.23$\pm$1.16 & 83.80$\pm$1.86 & 83.56$\pm$1.54 & 95.52$\pm$0.26 & 95.44$\pm$0.43 & \underline{95.03$\pm$0.46} & 86..90$\pm$1.06 & 87.06$\pm$1.56 & \underline{86.58$\pm$1.42} \\ 
			UEAF & 82.74$\pm$1.44 & 75.69$\pm$2.00 & 72.11$\pm$1.91 & 93.92$\pm$0.28 & 87.45$\pm$0.51 & 88.87$\pm$0.49 & 85.91$\pm$0.82 & 78.71$\pm$0.59 & 75.85$\pm$0.65 \\ 
   			IMSR & 82.25$\pm$1.14 & 82.11$\pm$1.25 & 82.13$\pm$0.53 & 94.80$\pm$0.49 & 94.76$\pm$0.39 & 94.35$\pm$0.35 & 85.19$\pm$0.95 & 85.51$\pm$0.92 & 85.27$\pm$0.48  \\ 
   			OS-LF-IMVC  & 90.19$\pm$1.33 & 80.70$\pm$1.62 & 56.11$\pm$2.00 & 96.68$\pm$0.34 & 91.78$\pm$0.55 & 80.57$\pm$0.73 & 91.29$\pm$0.99 & 81.49$\pm$1.13 & 58.53$\pm$1.77  \\ 
   			HCP-IMSC & \underline{91.65$\pm$0.55} & \underline{89.89$\pm$0.68} & \underline{85.02$\pm$3.48} & \underline{97.42$\pm$0.20} & \underline{96.73$\pm$0.47} & 93.86$\pm$2.33 & \underline{92.95$\pm$0.47} & \underline{91.51$\pm$0.56} & 86.56$\pm$3.87 \\ 
			Ours	&\textbf{93.03$\pm$0.36}	&\textbf{91.53$\pm$0.74}	&\textbf{90.89$\pm$1.15}	&\textbf{98.54$\pm$0.07}	&\textbf{98.19$\pm$0.11}	&\textbf{97.72$\pm$0.19}	&\textbf{94.78$\pm$0.33}	&\textbf{93.60$\pm$0.44}	&\textbf{92.88$\pm$0.77}\\
			\bottomrule
		\end{tabular}
	}
	\label{tab:res5}

\end{table*}
\begin{figure*}[!t]
	\centering
	\subfloat[Missing views on NH\_face and Handwritten databases]{
		\includegraphics[width=0.15\textwidth]{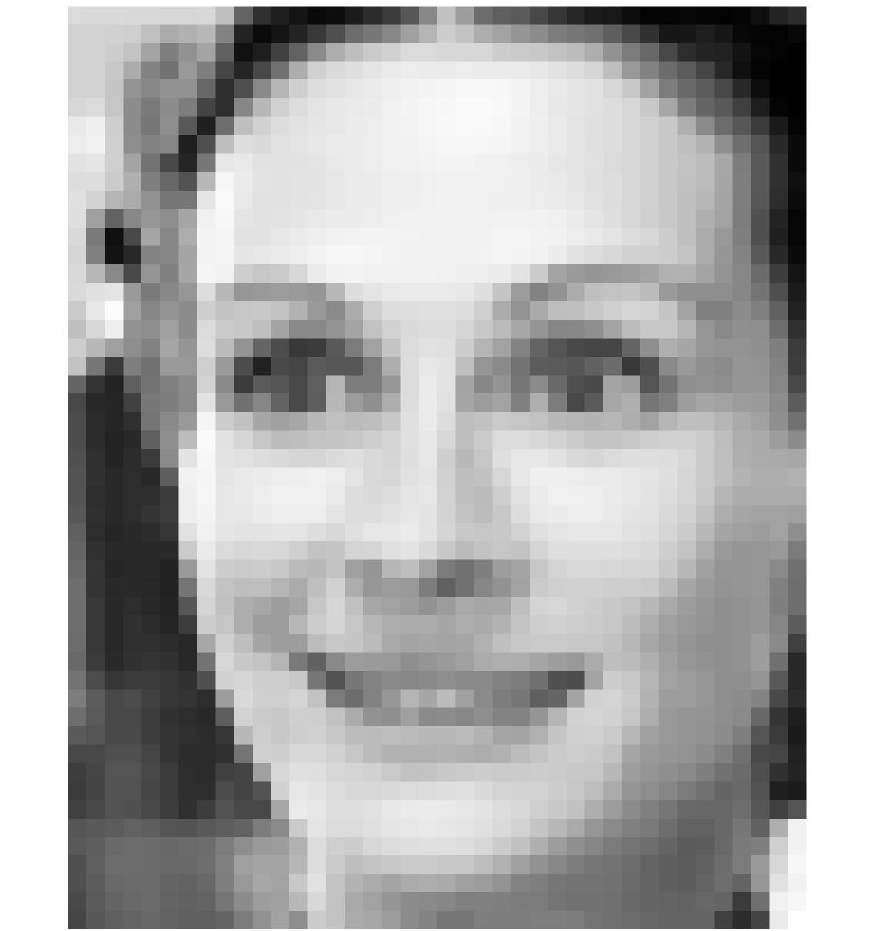}
		\includegraphics[width=0.15\textwidth]{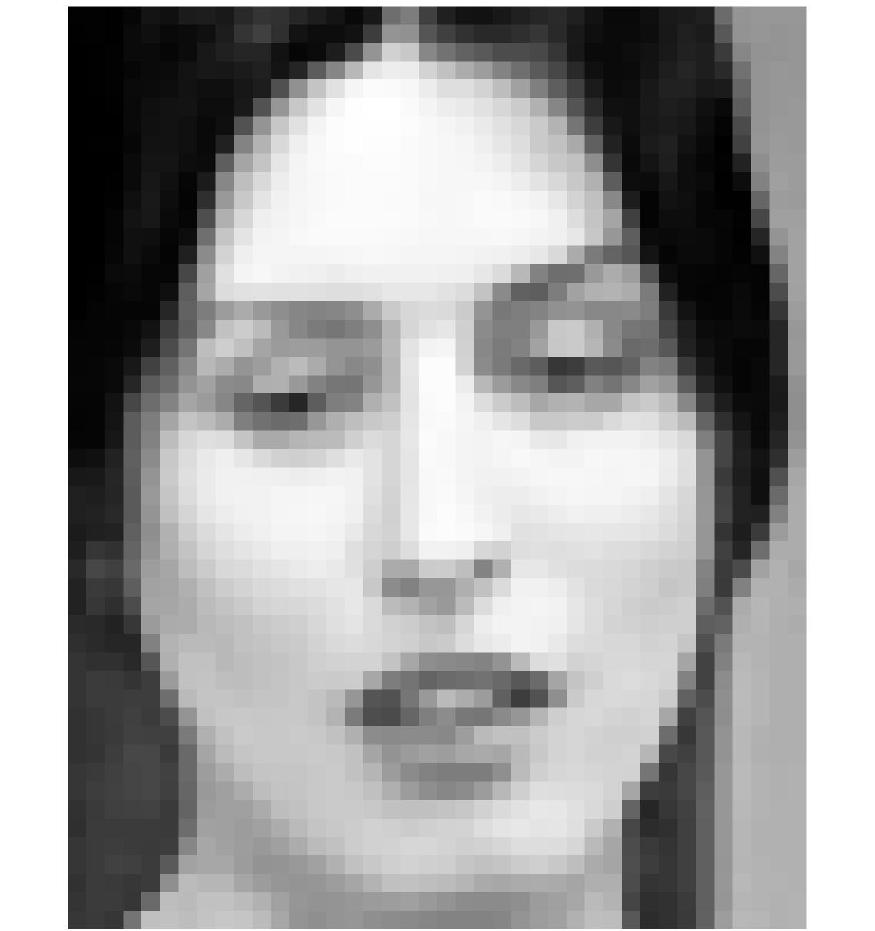}
		\includegraphics[width=0.15\textwidth]{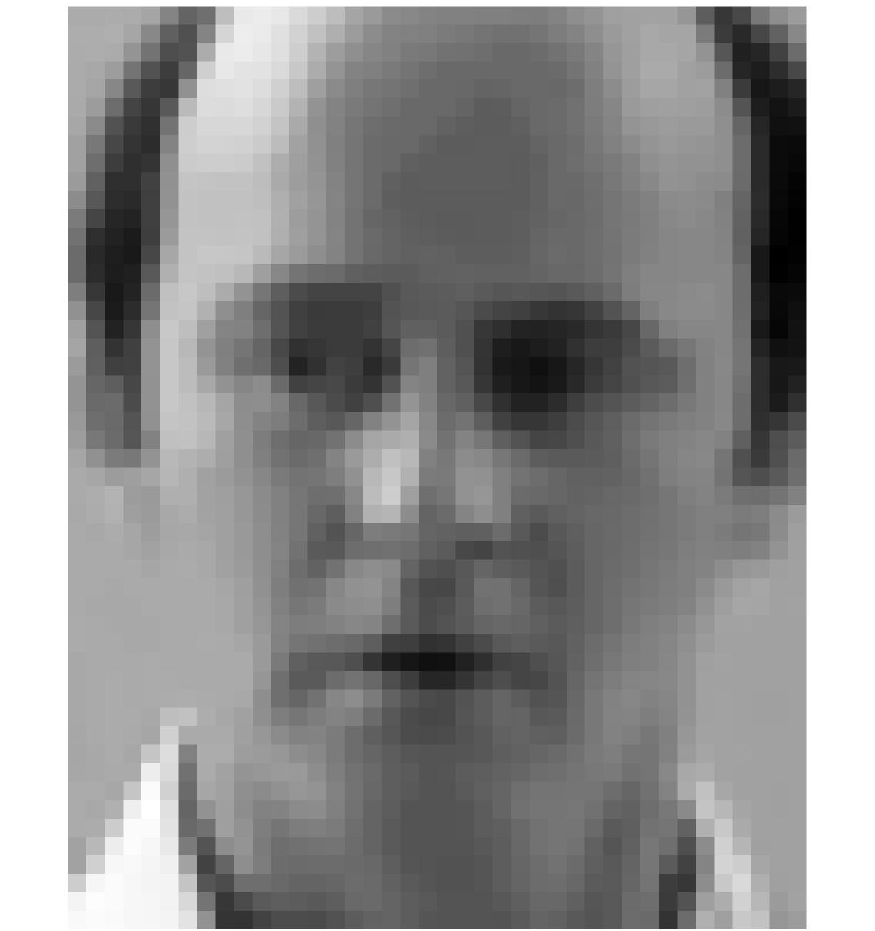}
		\includegraphics[width=0.15\textwidth]{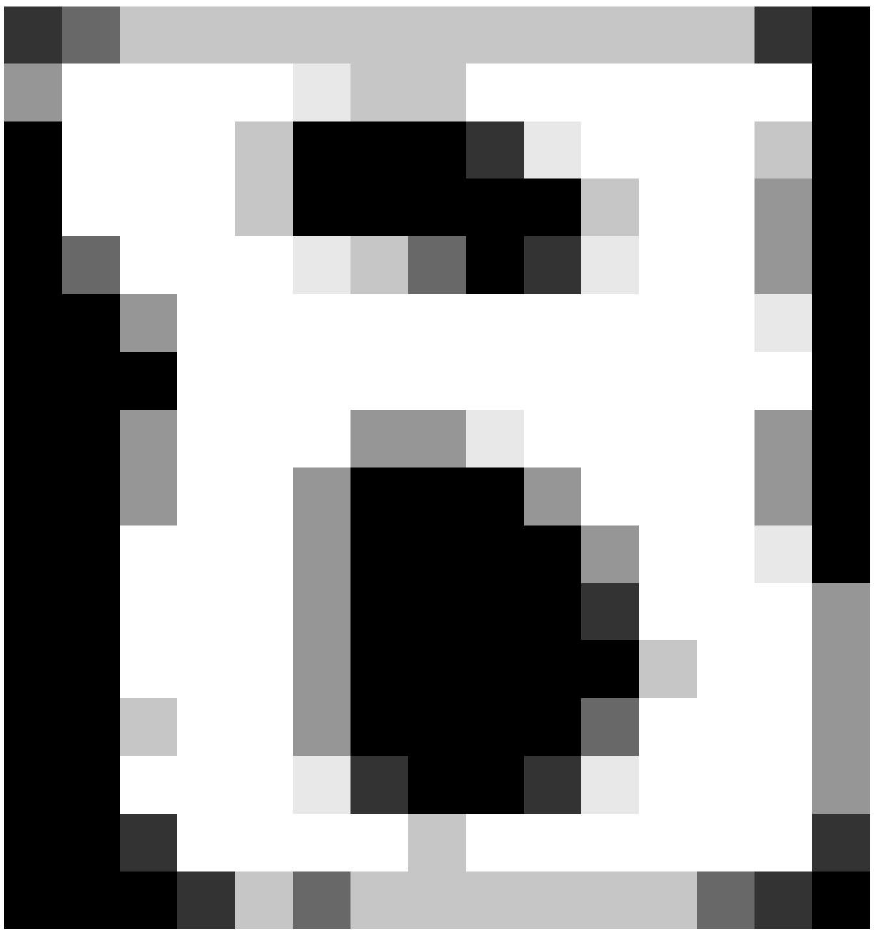}
		\includegraphics[width=0.15\textwidth]{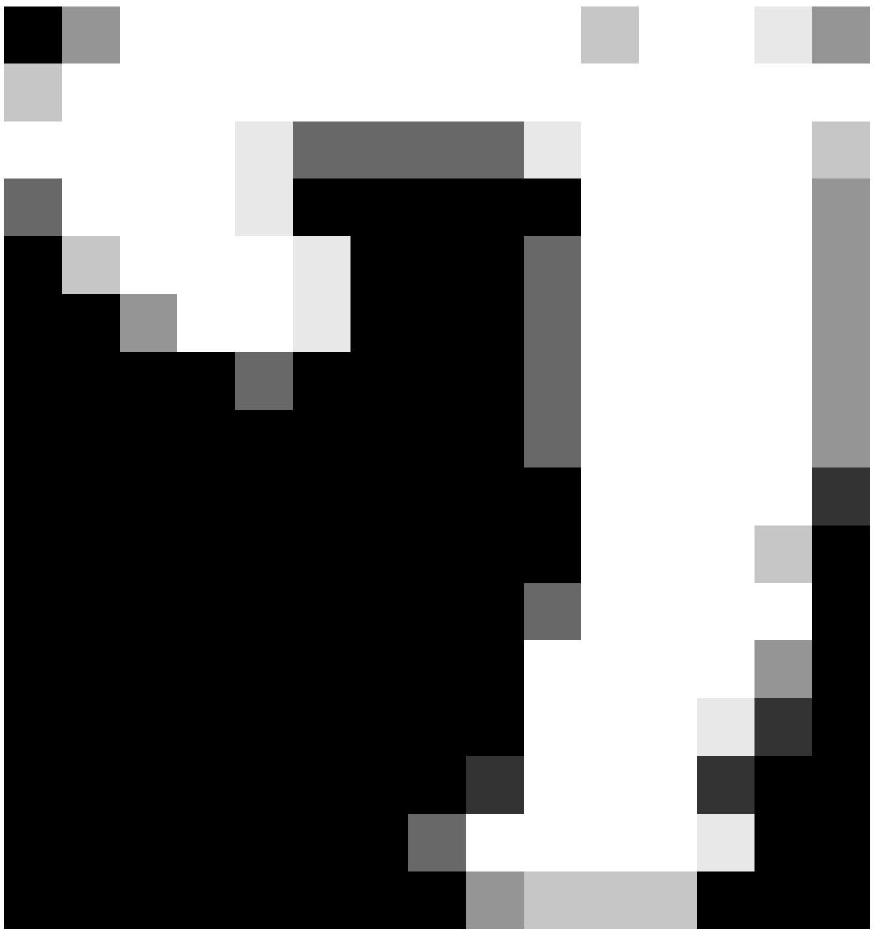}
		\includegraphics[width=0.15\textwidth]{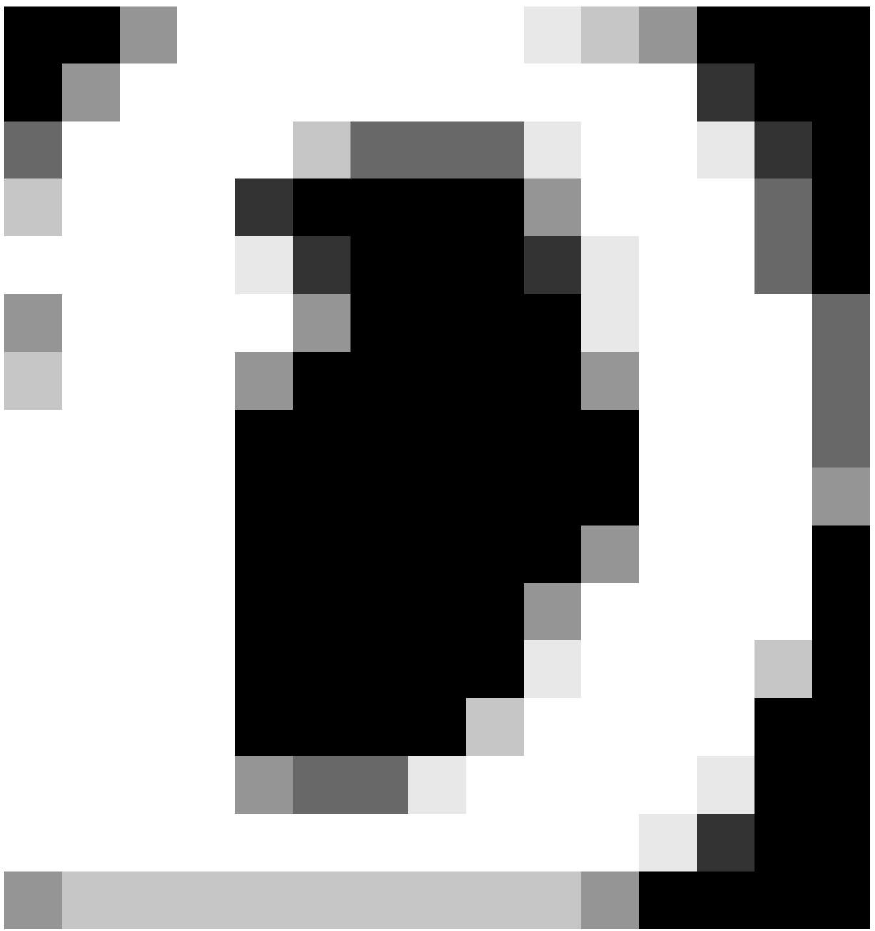}
	}
	\vspace{0.3cm}
	\quad
	\subfloat[Recovered views on NH\_face and Handwritten databases]{
		\includegraphics[width=0.15\textwidth]{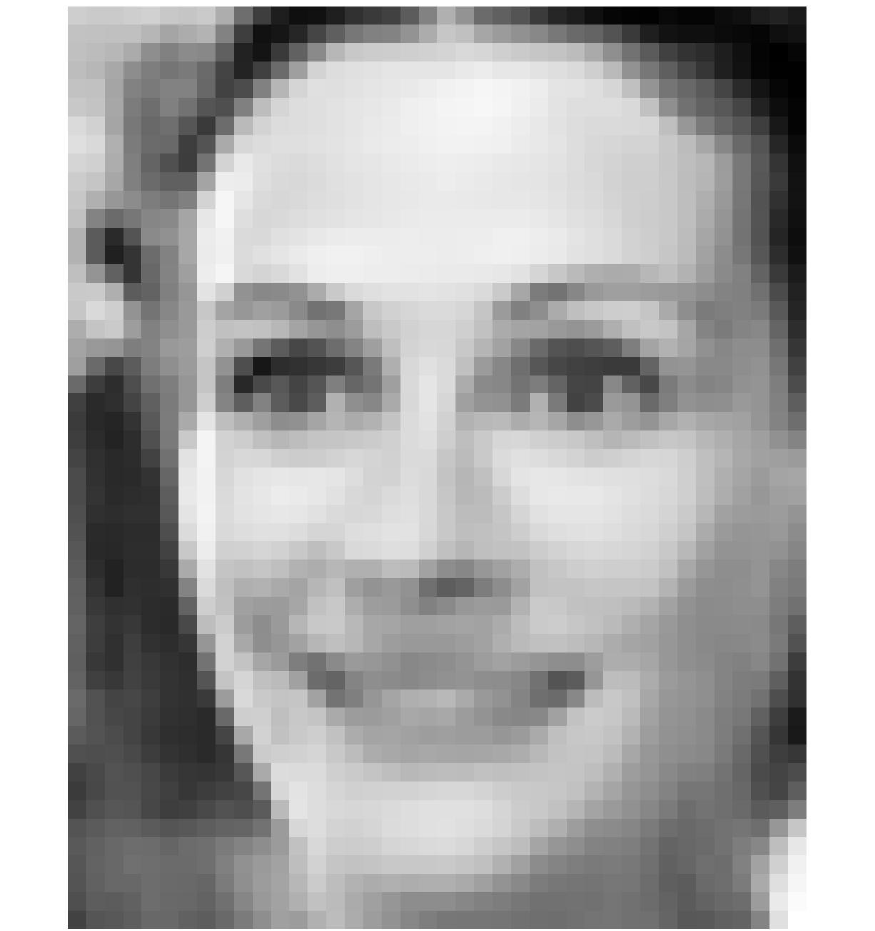}
		\includegraphics[width=0.15\textwidth]{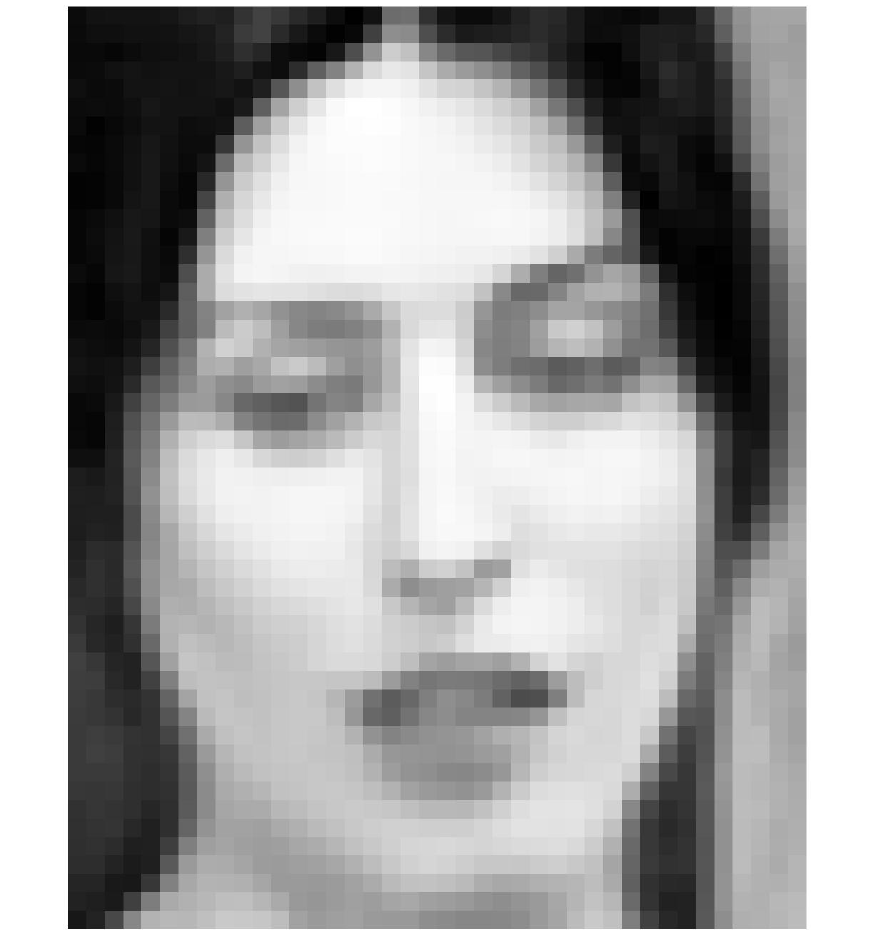}
		\includegraphics[width=0.15\textwidth]{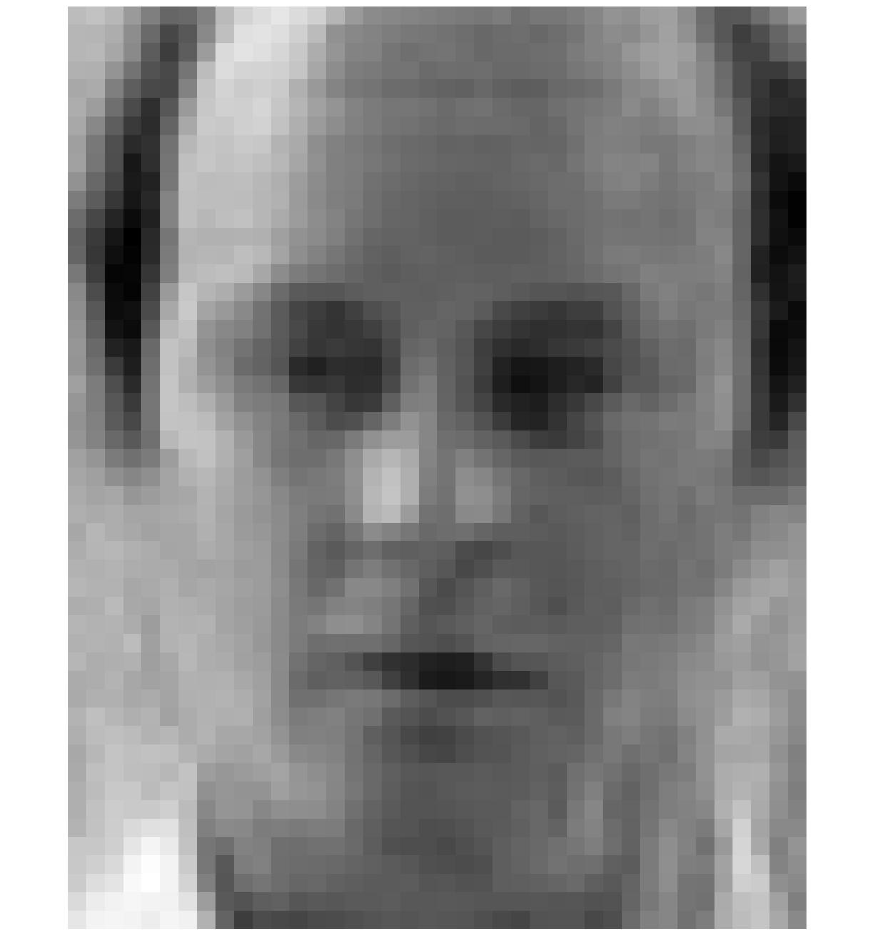}
		\includegraphics[width=0.15\textwidth]{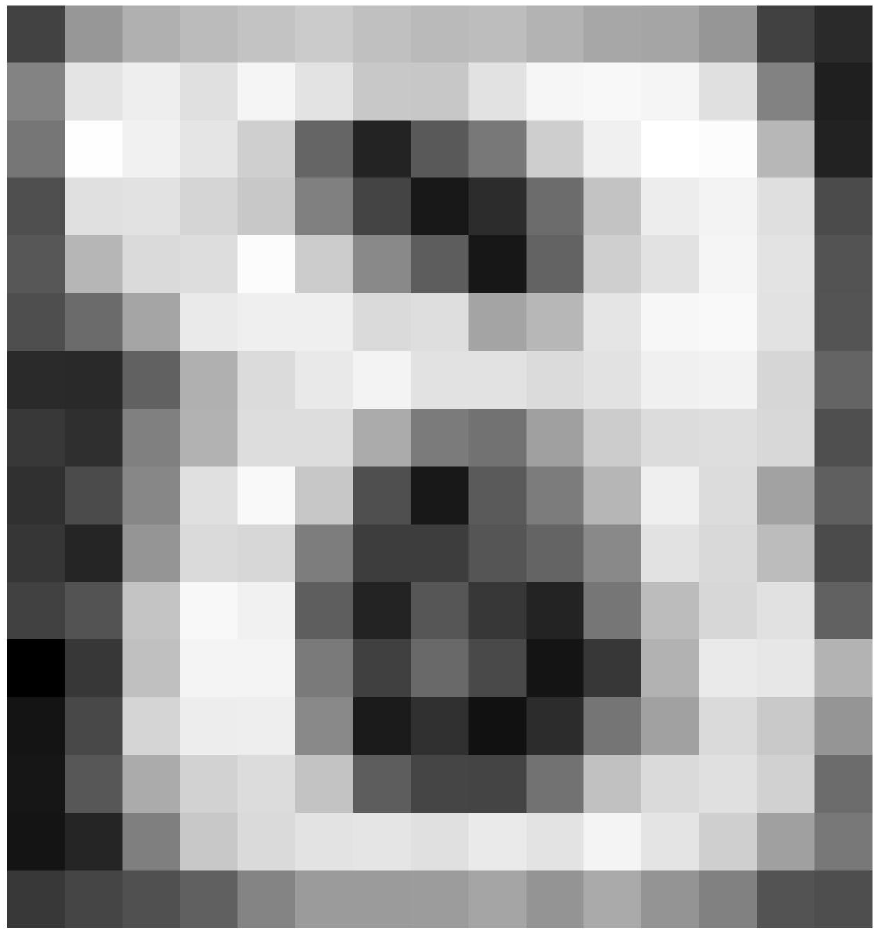}
		\includegraphics[width=0.15\textwidth]{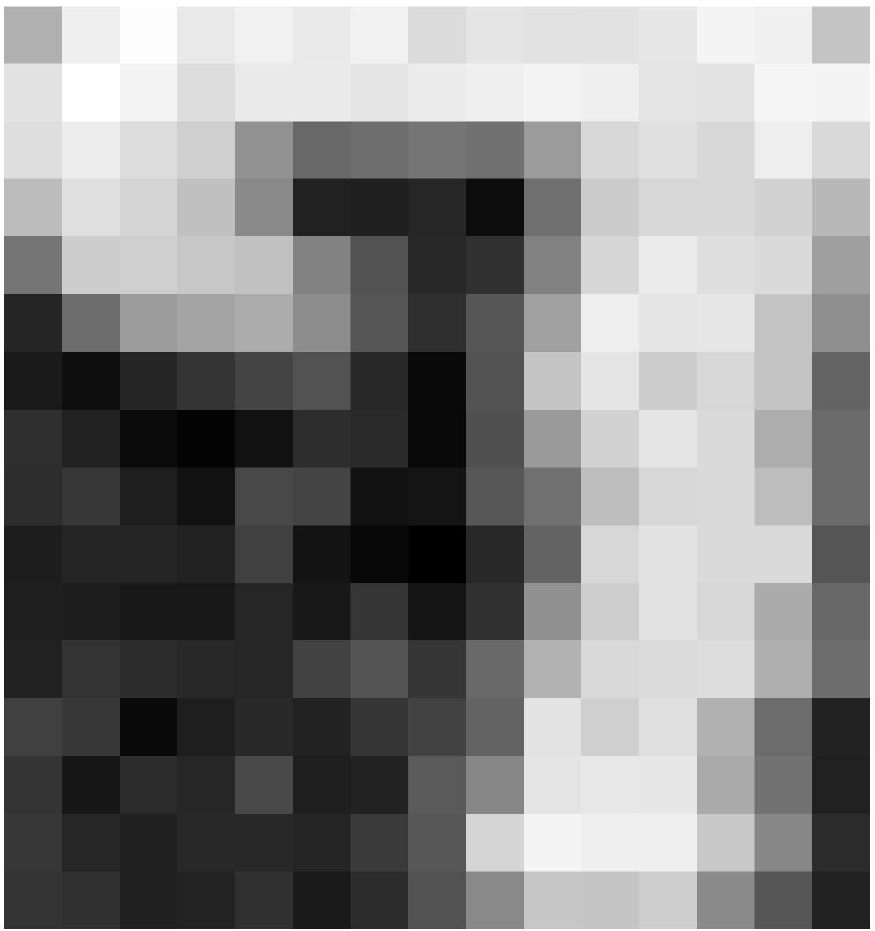}
		\includegraphics[width=0.15\textwidth]{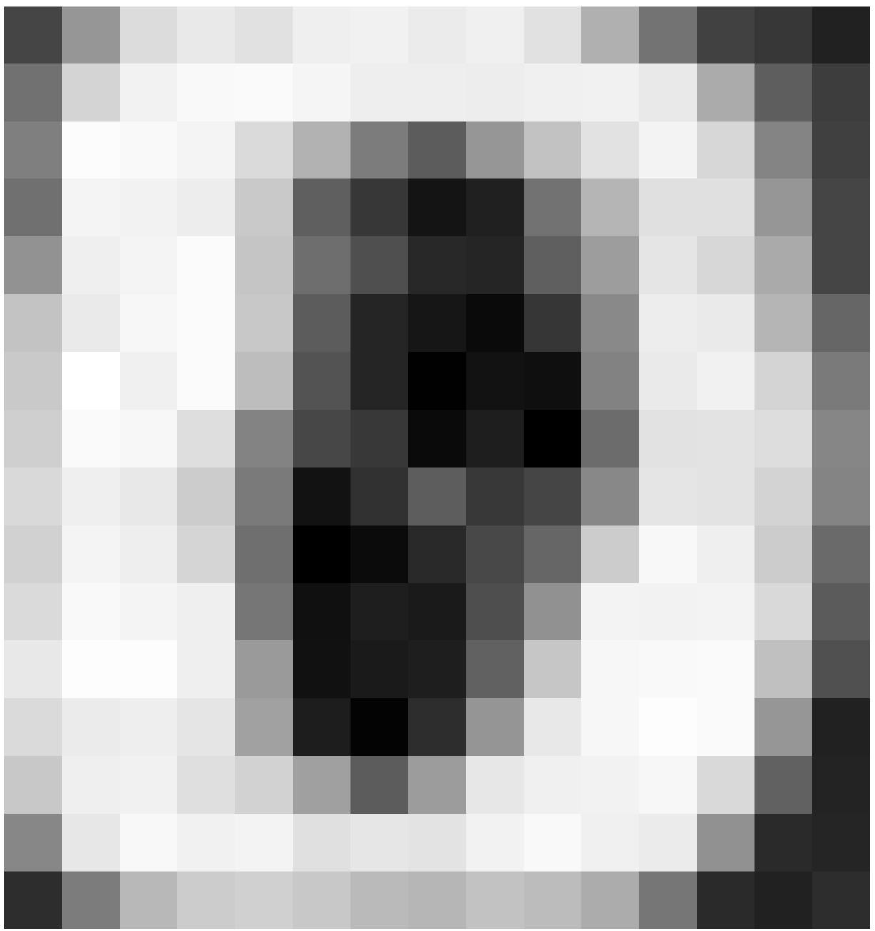}
	}
	\caption{The visual example pairs about missing views and their restoration results. The (a) shows the missing views sampling from the NH\_face and Handwritten databases; The (b) denotes corresponding views recovered by our RecFormer.}
	\label{fig:recovery}
\end{figure*}

\textbf{Evaluation}: Following \cite{yuan2021adaptive,wen2019unified,liu2020efficient}, we still select the clustering accuracy (ACC), normalized mutual information (NMI), and purity as our three metrics to evaluate these methods. The higher the values of the three metrics, the better the clustering performance. Besides, all comparison methods are performed multiple times to reduce randomness and their parameters are set as suggested in their papers or codes for a fair comparison.

\textbf{Implementation details}: In our experiments, the two transformer-style cross-view en-decoders are set as one layer with 4 heads. The learning rate is $0.001$ and we select the Adaptive Moment Estimation (Adam) as our optimizer. For all databases, we set 50 epochs for each of the two stages and batch size is 128. Our method is performed on a personal computer with an Intel 10700k CPU, RTX2080s GPU, Ubuntu 20.04, and python 3.9.13. Our code\footnote{Code: \url{https://github.com/justsmart/Recformer-mindspore}. } is implemented by Mindspore 2.0.0 and Pytorch 1.13.0. 

\subsection{Experimental results and analysis}

As shown in Tables \ref{tab:res1}-\ref{tab:res5}, the values of ACC, NMI, and Purity are listed and corresponding standard deviations are given after the sign '$\pm$'. We report the twelve methods on each dataset with different incomplete rates or paired rates in these tables and mask the best or second-best results in bold or underline. Looking closely at these data, we can easily get a few points: 
\begin{itemize}
\item Other methods designed for the IMC task achieve significantly better performance than baseline methods (BSV and Concat) that cannot handle missing views. This confirms the necessity of considering missing views in the design of multi-view clustering models.
\item Our approach shines brightly, beating other state-of-the-art methods in almost all metrics. For instance, our RecFormer exceeds the second-best DAIMC in the ACC metric by approximately 10, 11, and 11 percentage points on NH\_face dataset with three different incomplete rates, respectively. The good performance of our method on the clustering demonstrates that the high-level semantic representation extracted by it is effective and solid. 
\item Comparing all results horizontally, we conclude that the harm of missing views to multi-view learning is definite and the higher missing rates lead to worse learning outcomes in most cases, which is intuitive and comprehensible. Besides, different IMC methods have different immunity to data incompleteness, such as, UEAF, HCP-IMSC, and our approach are relatively insensitive to the missing views due to its powerful view recovery capability.
\end{itemize}

In Figure \ref{fig:recovery}, we give the six visual recovery results by reshaping the pixel feature in the first view of Handwritten database and the second view of NH\_face database. The first row is the raw missing views, and the second row denotes the recovery results of our method. As we can see, our RecFormer has an amazing recovery effect on missing views, which is very beneficial for the subsequent generation of approximately complete graph. 
\begin{figure*}[t!]
	\centering
	\subfloat[BSV]{
		\includegraphics[width=0.24\textwidth]{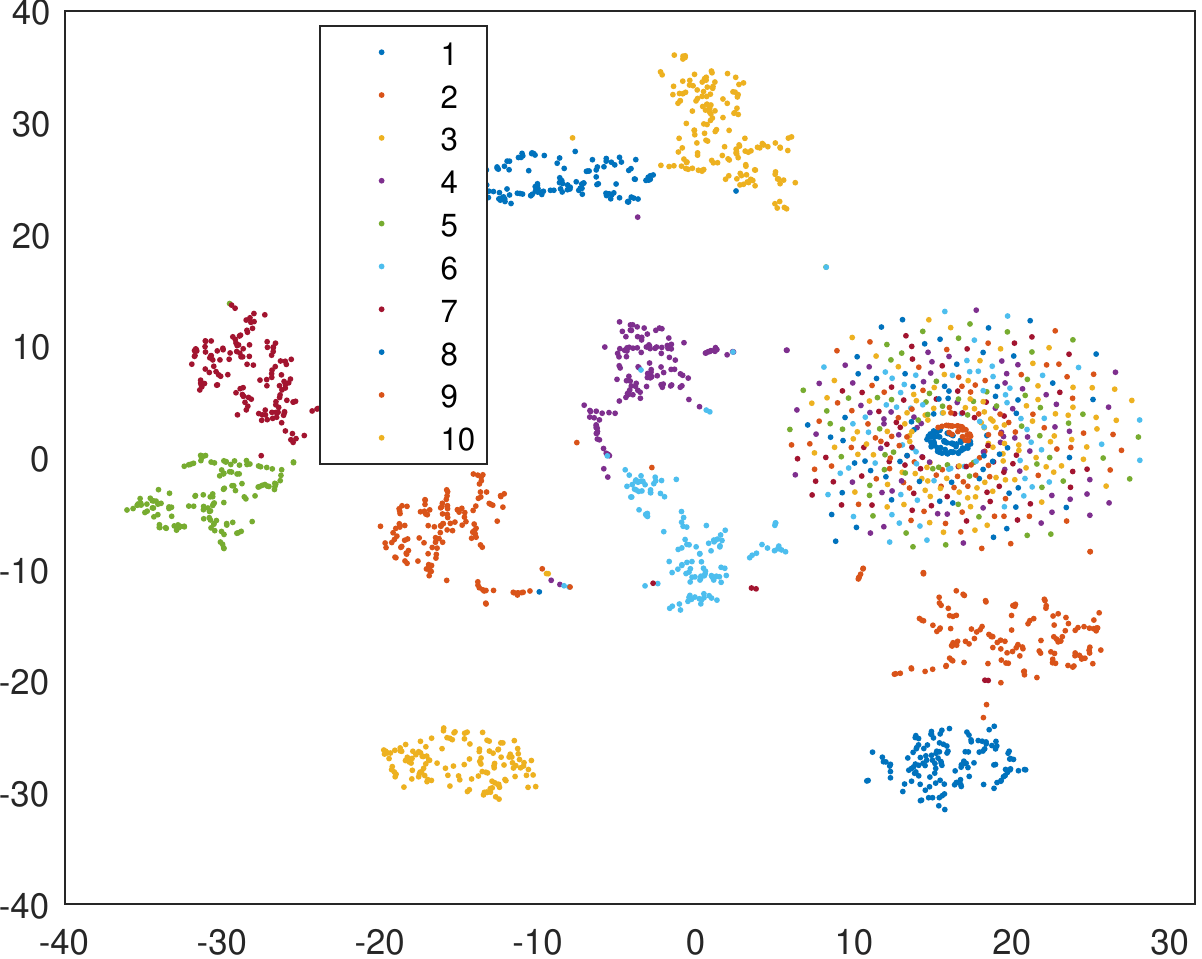}
	}
	\subfloat[Concat]{
		\includegraphics[width=0.24\textwidth]{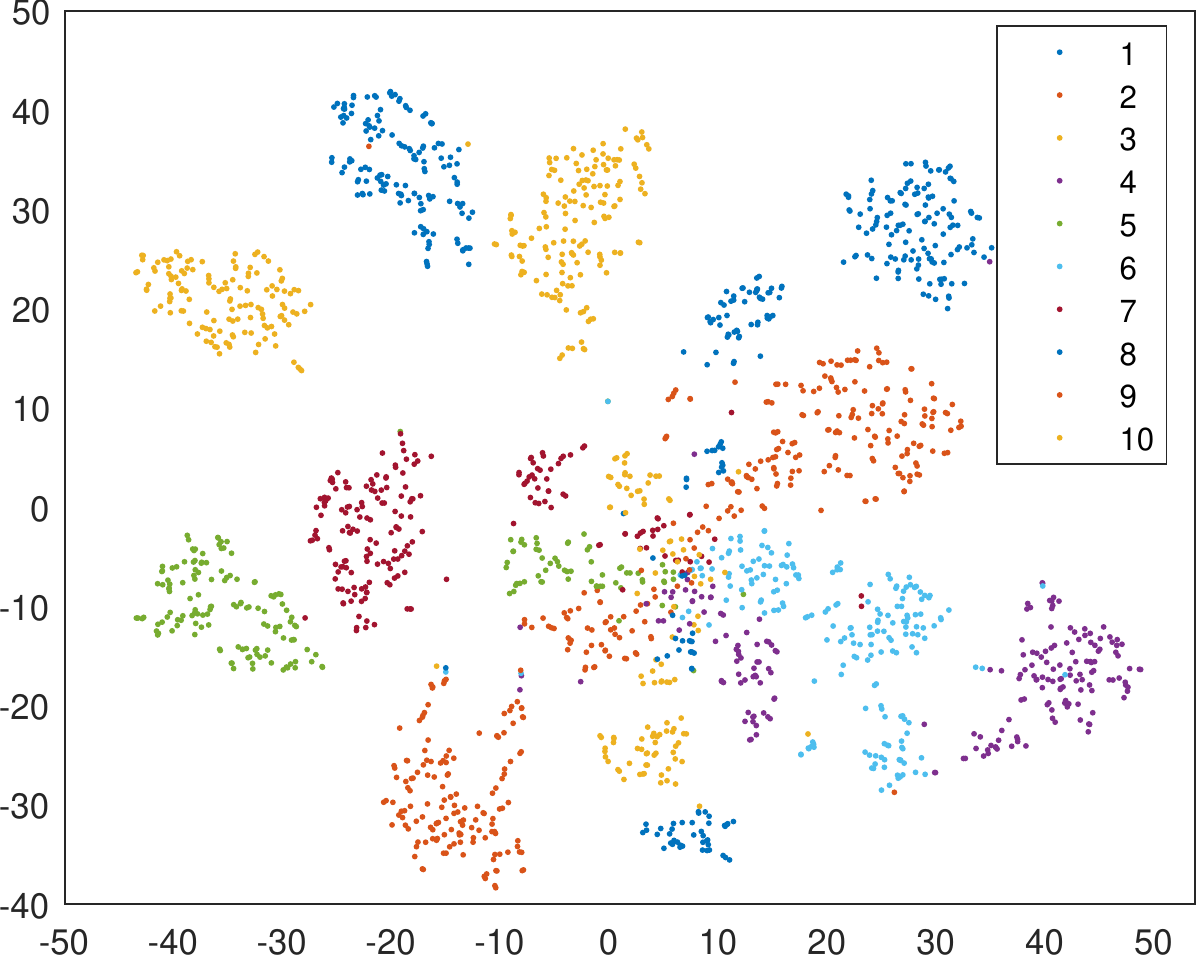}
	}
	\subfloat[MIC]{
		\includegraphics[width=0.24\textwidth]{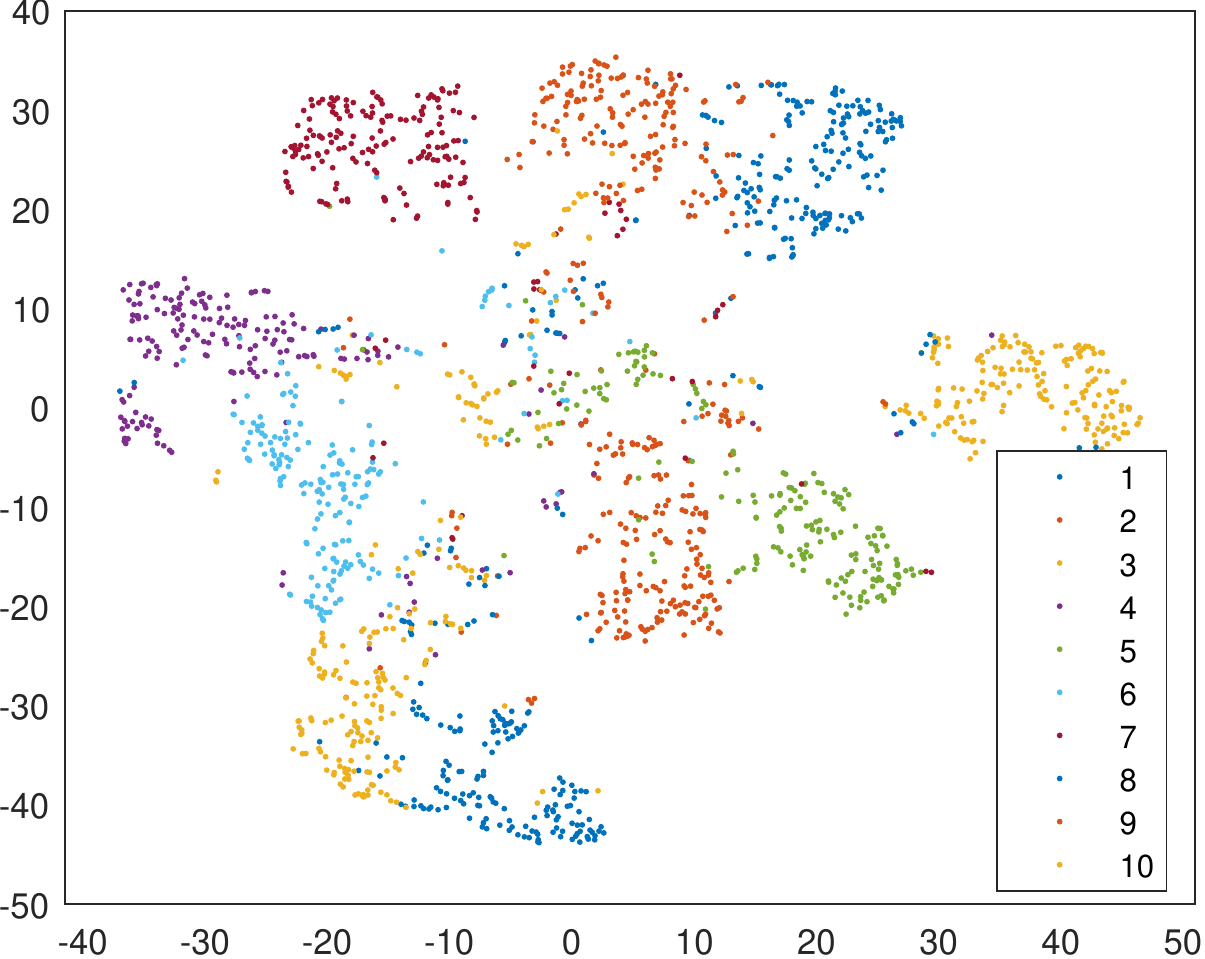}
	}
	\subfloat[DAIMC]{
		\includegraphics[width=0.24\textwidth]{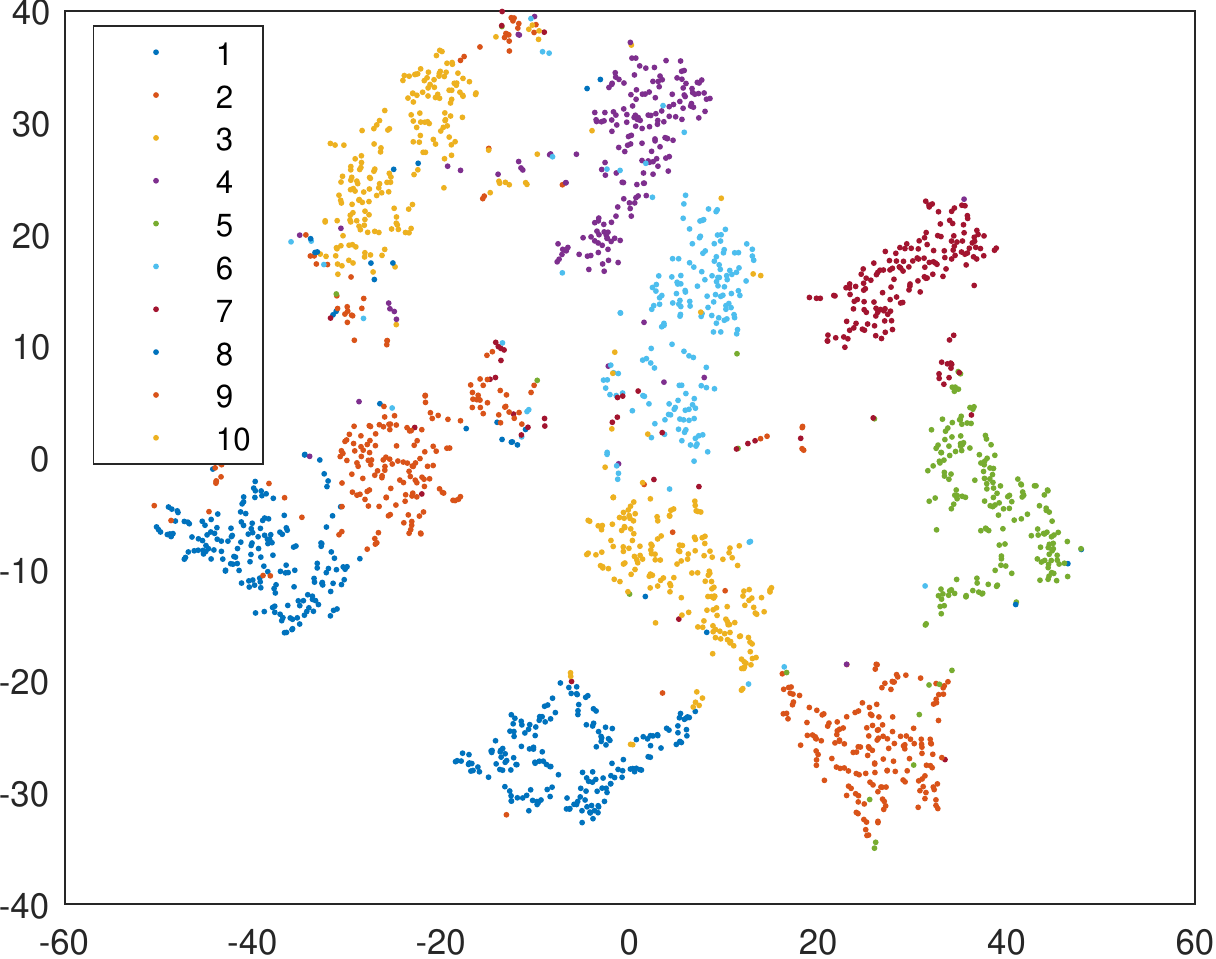}
	}
	\quad
	\subfloat[OMVC]{
		\includegraphics[width=0.24\textwidth]{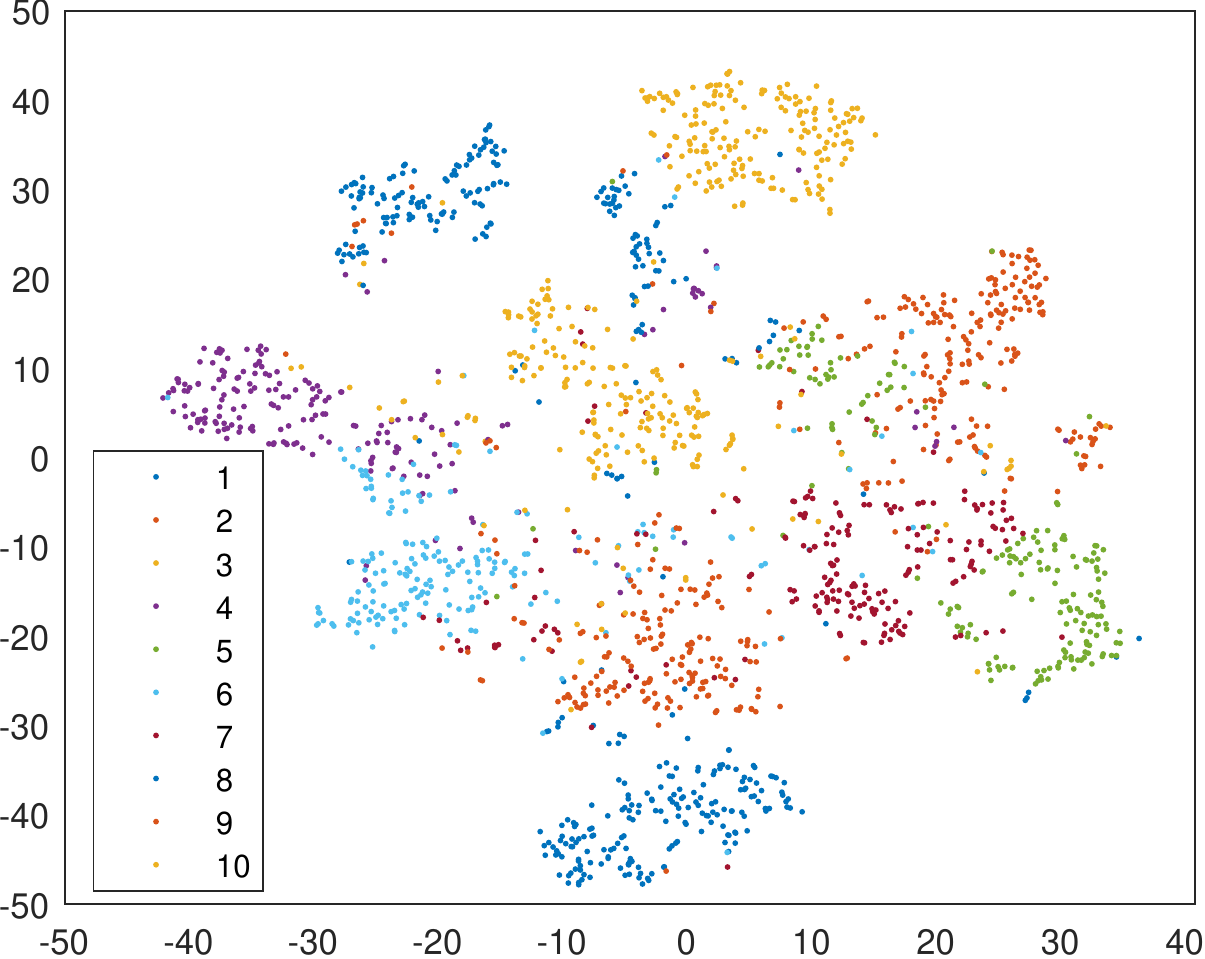}
	}
	\subfloat[MKKM-IK-MKC]{
		\includegraphics[width=0.24\textwidth]{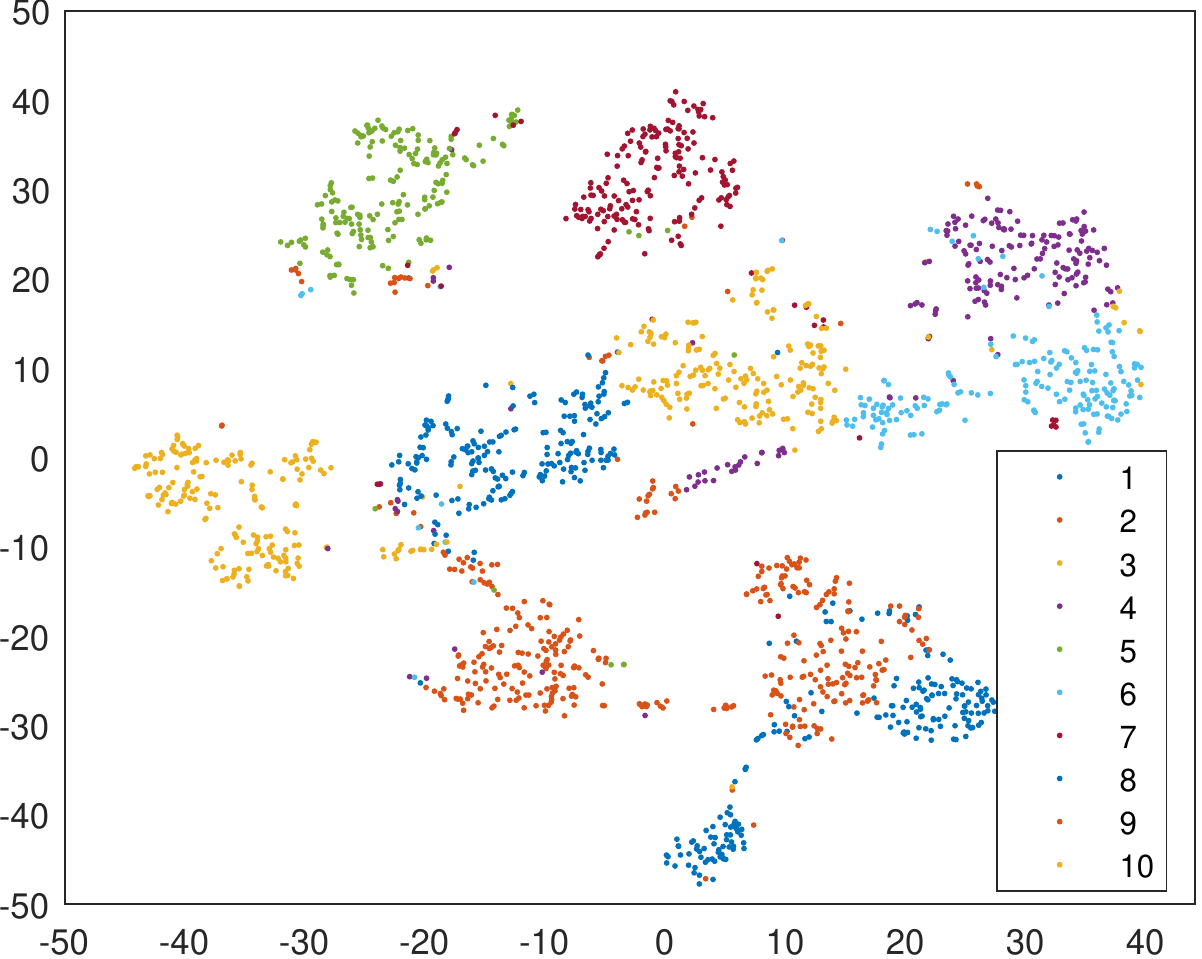}
	}
	\subfloat[UEAF]{
		\includegraphics[width=0.24\textwidth]{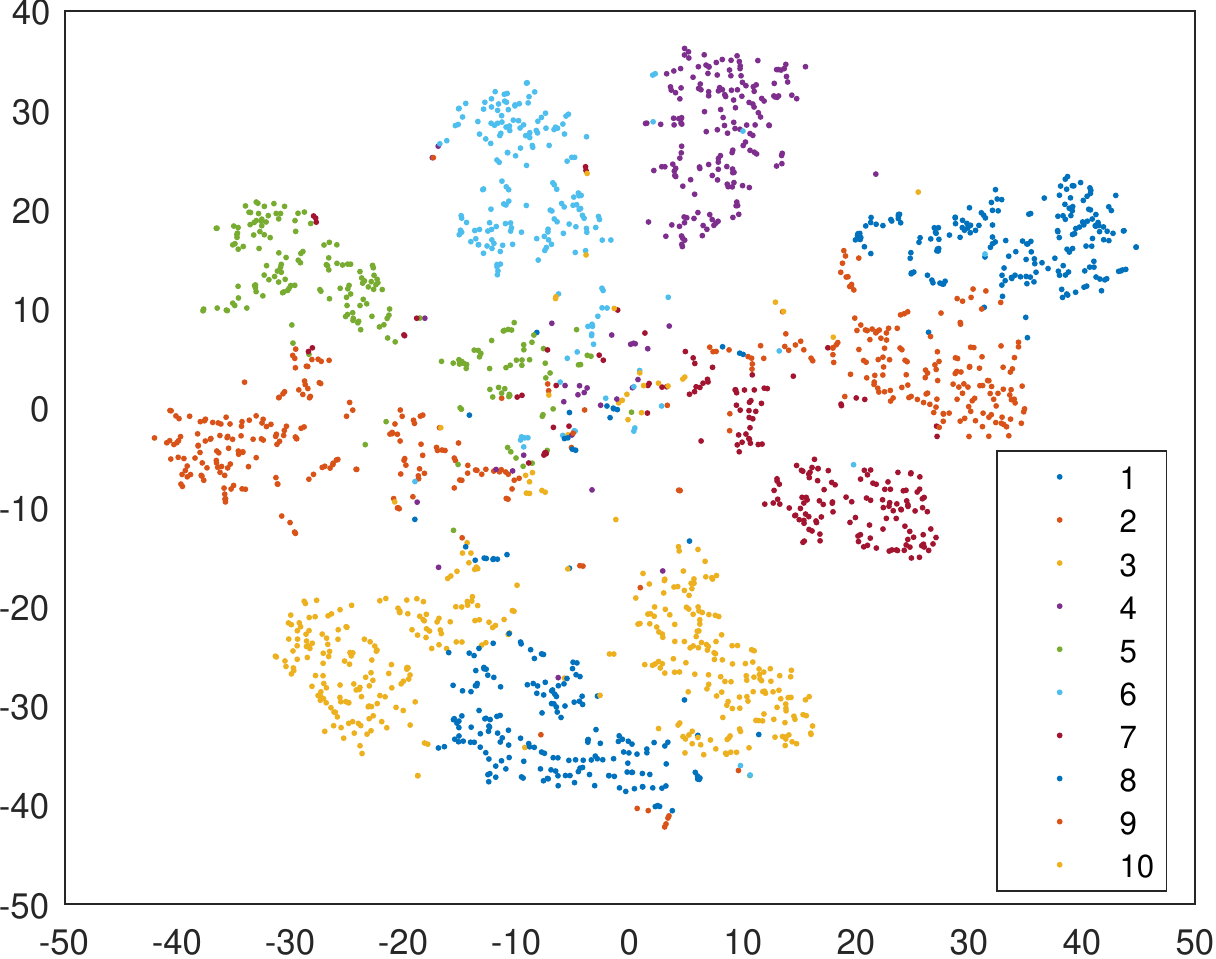}
	}
	\subfloat[IMSR]{
			\includegraphics[width=0.24\textwidth]{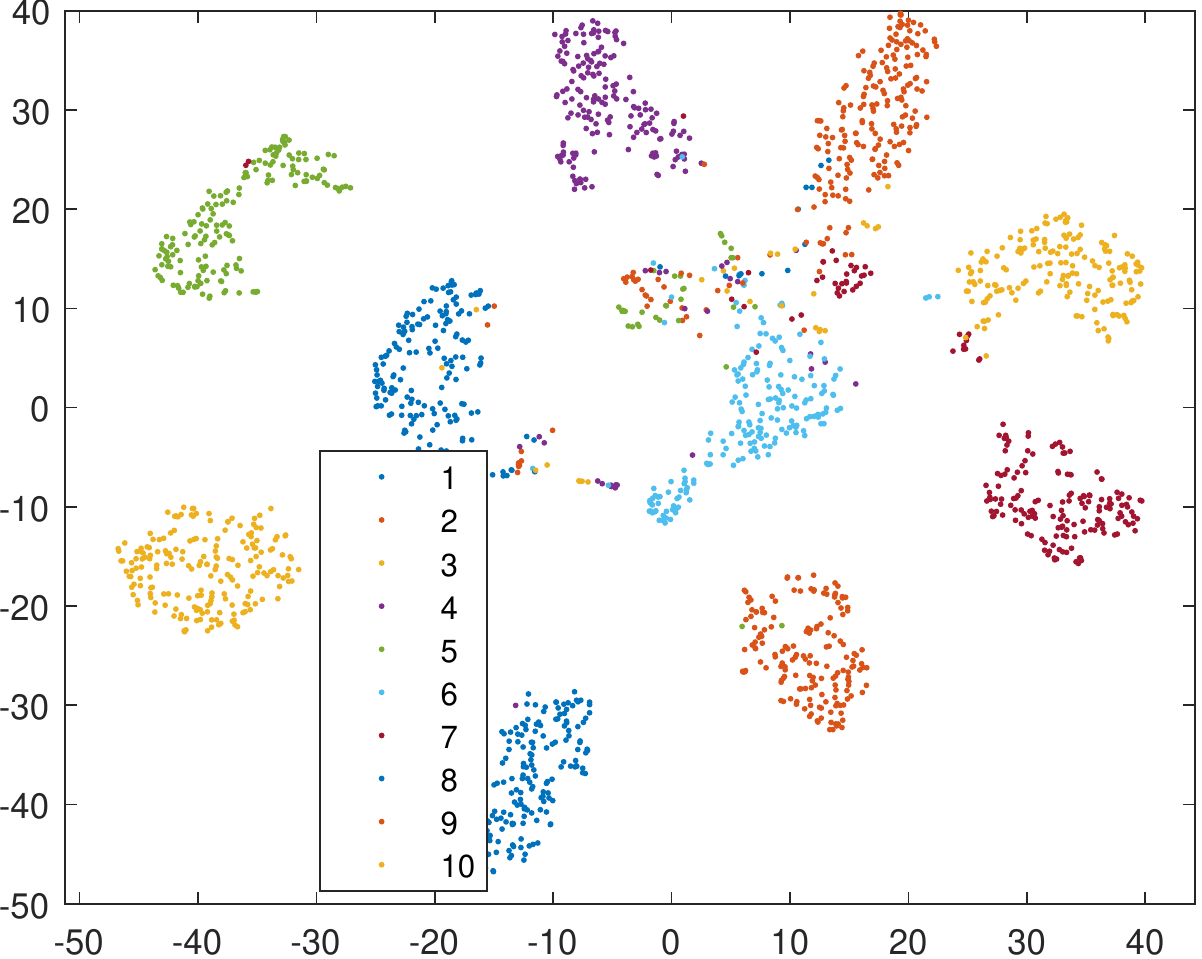}
		}
	\quad
	\subfloat[HCP-IMSC]{
			\includegraphics[width=0.24\textwidth]{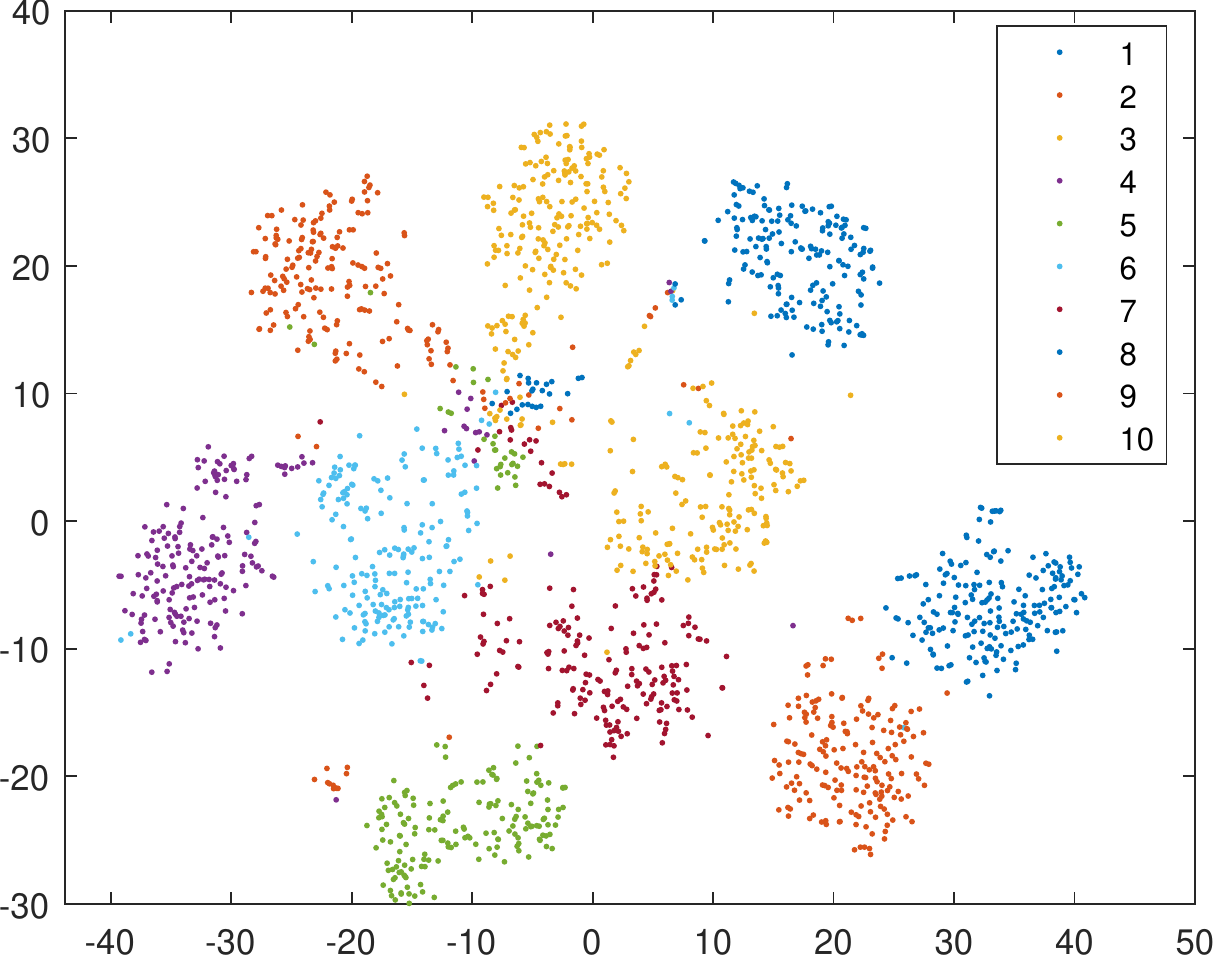}
		}
	\subfloat[RecFormer]{
		\includegraphics[width=0.24\textwidth]{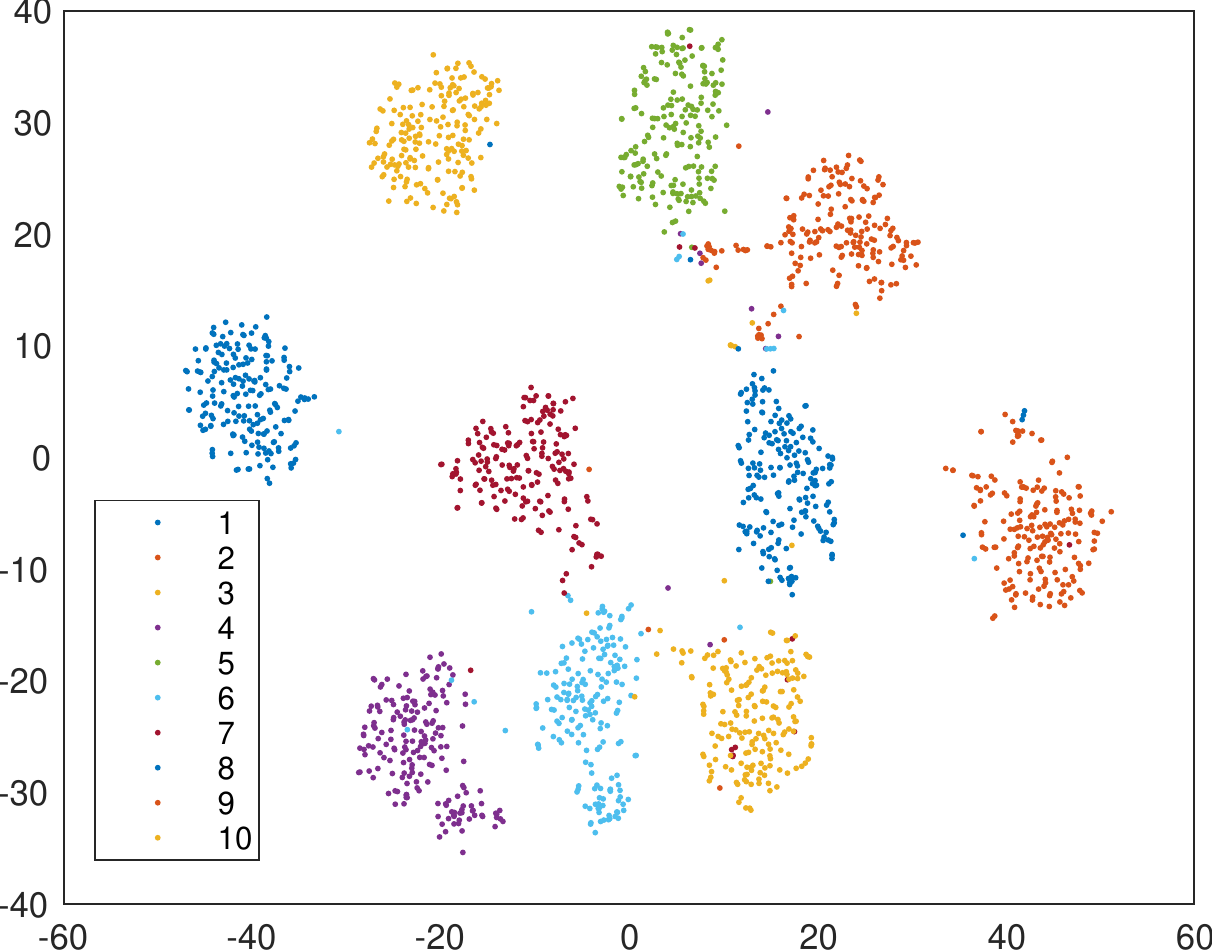}
	}
	\caption{Feature space visualization of final clustering representations of different methods via t-SNE on the Handwritten dataset with a 30\% incomplete rate.}
	\label{fig.tsne}

\end{figure*}
\subsection{Ablation study}
To confirm the effectiveness of each component of our RecFormer, we respectively remove the cross-view en-decoder module leaving only the FC layers, recurrent graph constraint, missing indicator matrix, and second stage to form new models. Results of these methods on Handwritten and NH\_face databases with a 50\% missing ratio are shown in Table \ref{table:ablation}. To be specific, $\mathcal{B}$ is the backbone with only linear layers; $\mathcal{C}$ denotes the cross-view en-decoder module; $\mathcal{G}$ represents the current graph constraint; and incomplete mask means the missing prior matrix introduced in our proposed RecFormer. In addition, we try to remove Stage 2 and use the fusion feature $\bar{Z}$ in Stage 1 to perform clustering (keep the same training epochs). From the table, it is clear that the full version of the RecFormer achieves the best performance. Furthermore, we find that the introduction of a missing prior matrix is crucial for the recovery phase of RecFormer, which well avoids the negative influence of missing data. And we can see that Stage 2 is beneficial to performance improvement, thanks to the introduction of recovered data in Stage 1.

\begin{table}[h]
	
	\renewcommand{\arraystretch}{1.3}
	\caption{The ablation experiments on two datasets with a 50\% missing ratio. $\mathcal{B}$ is the backbone with only linear layers; $\mathcal{C}$ denotes the cross-view en-decoder module; $\mathcal{G}$ represents the current graph constraint; and incomplete mask means the missing prior matrix introduced in RecFormer.}
	\label{table:ablation}
	\centering
	\begin{tabularx}{\linewidth}{c|XXX|XXX}
		\toprule[1pt]
		\multirow{2}{*}{method}  & \multicolumn{3}{c|}{Handwritten} & \multicolumn{3}{c}{NH\_face}\\
		& ACC   & NMI   & PUR & ACC   & NMI   & PUR  \\
		\midrule
		$\mathcal{B}$
		&85.80	&77.43	&85.80	&49.61	&27.70	&54.29\\	
		$\mathcal{B}+\mathcal{C}$
		&89.85	&80.02	&89.85	&82.81	&77.24	&84.70\\
		$\mathcal{B}+\mathcal{G}$
		&88.05	&77.99	&88.05	&79.98	&73.05	&82.85\\
		$\bm{\mathcal{B}+\mathcal{C}+\mathcal{G}}$
		&\textbf{91.74}	&\textbf{83.39}	&\textbf{91.74}	&\textbf{95.40}	&\textbf{92.01} &\textbf{95.48}\\
		\textit{w/o} incomplete mask
		&55.40	&47.92	&55.50	&58.78	&48.62 &64.29\\
		\textit{w/o} Stage 2
		&89.94	&80.83	&89.94	&93.64	&90.02 &93.95\\
		\bottomrule[1pt]
	\end{tabularx}
	
\end{table}
\vspace{-0.2cm}
\subsection{T-SNE visualization results}

To visualize the clustering performance of these methods, in Figure \ref{fig.tsne}, we show the final clustering representation of different methods on the Handwritten database with a 30\% missing rate via the t-SNE \cite{van2008visualizing} technology. The results of OPIMC and OS-LF-IMVC are ignored in the figure because they obtain the clustering result directly without producing any clustering indicator matrix. Comparing all t-SNE results, it's easy to find that our RecFormer enjoys the best discrimination performance than other state-of-the-art methods.

\subsection{Convergence study}

\begin{figure}[t]
	\centering
	\subfloat[Handwritten]{
		\includegraphics[width=0.49\linewidth]{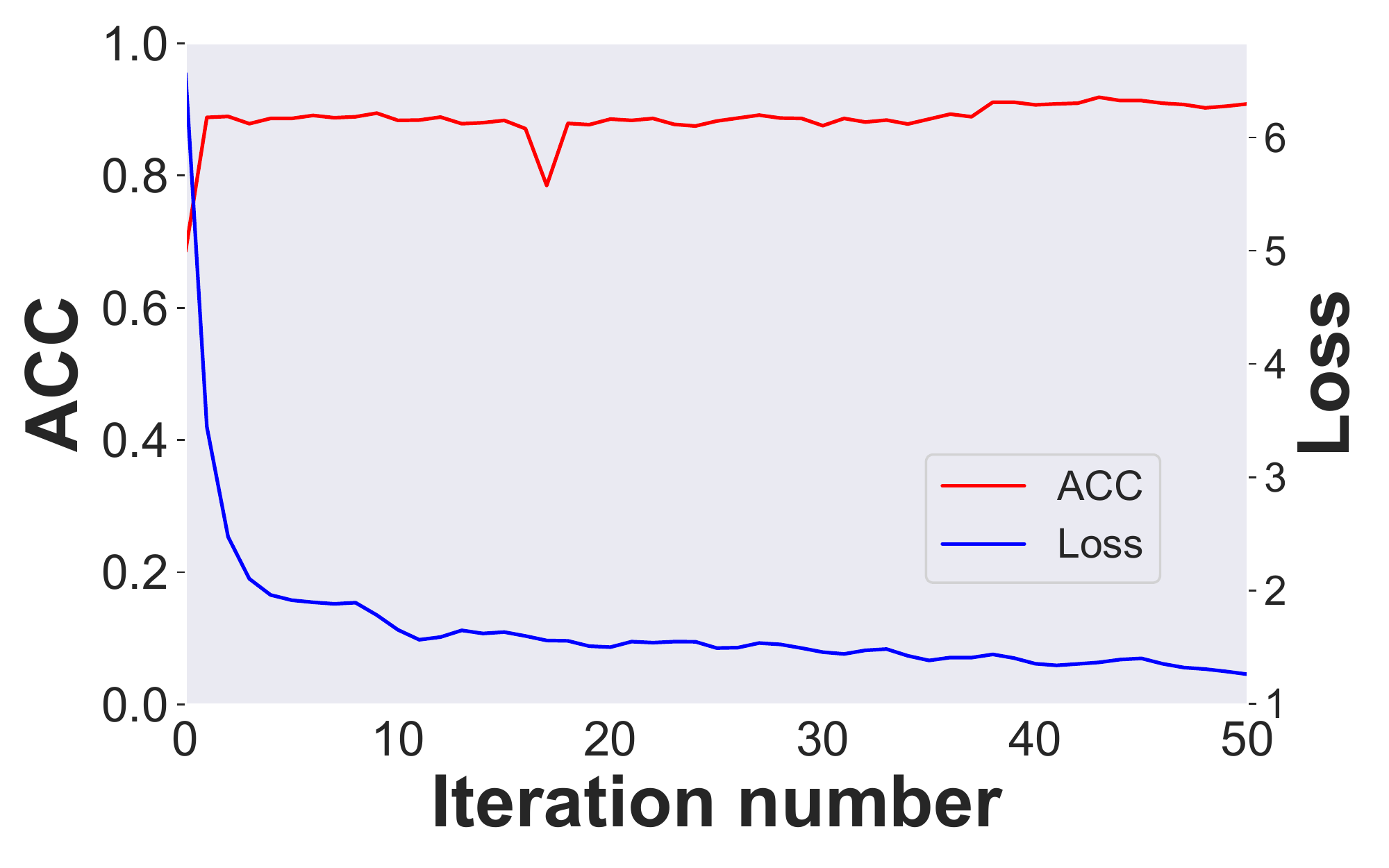}
	}
	\subfloat[Caltech7]{
		\includegraphics[width=0.49\linewidth]{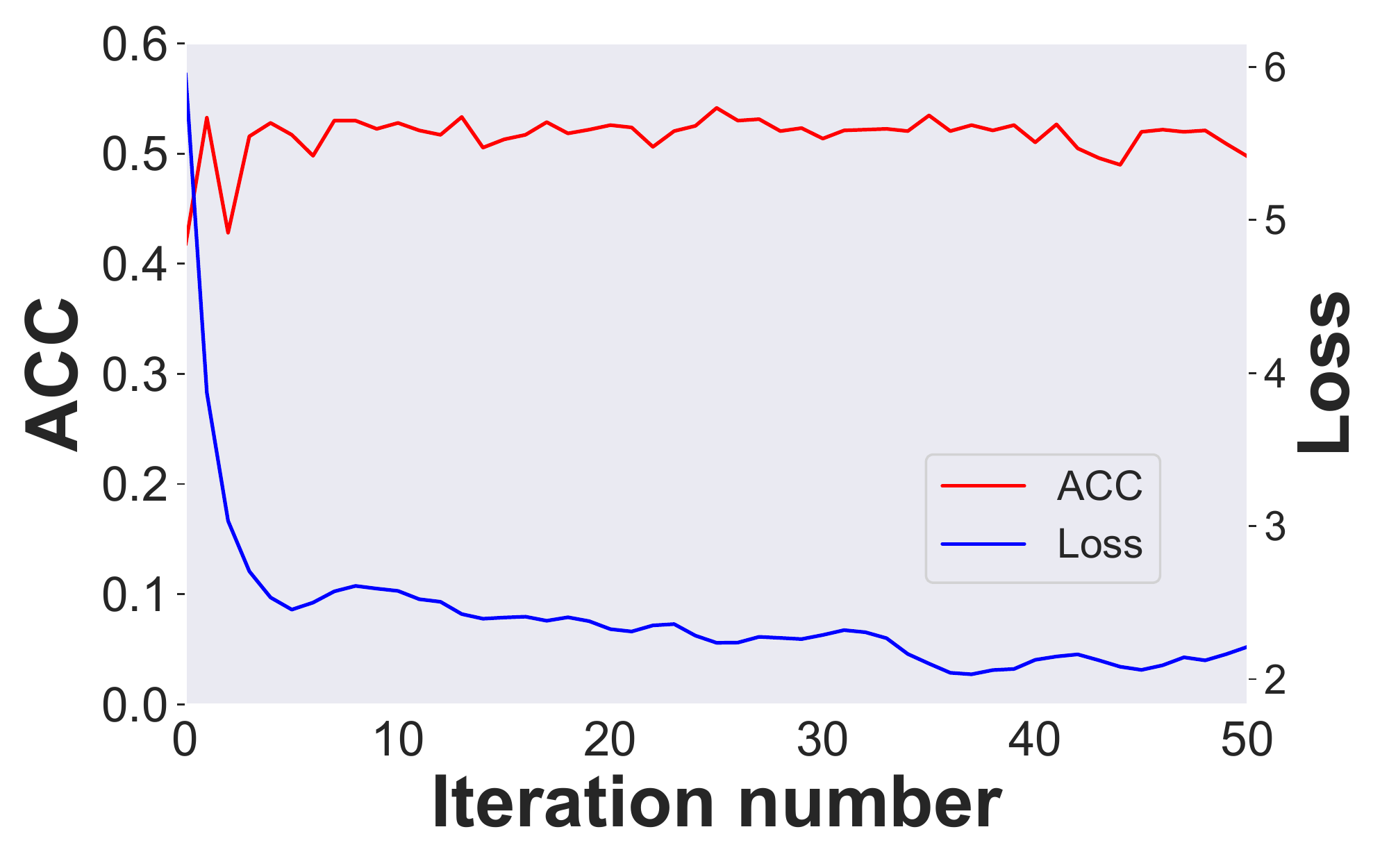}
	}
	\caption{The ACC and Loss curves on the Handwritten dataset and Caltech7 dataset with a 50\% missing rate in Stage 2.}
	\label{fig.conv}
	\vspace{-0.3cm}
\end{figure}

In order to study the convergence of our RecFormer, in Figure \ref{fig.conv}, we plot two ACC-Loss curves on the Handwritten dataset and Caltech7 dataset with a 50\% missing rate for Stage 2. From the figure, thanks to good data recovery in Stage 1, the loss value decreases steadily while the ACC keeps a slow upward trajectory, which demonstrates that our model has a good convergence. 

\subsection{Hyperparameters sensitivity study}
\begin{figure}[h]
\centering
\subfloat[Handwritten]{
	\includegraphics[width=0.49\linewidth]{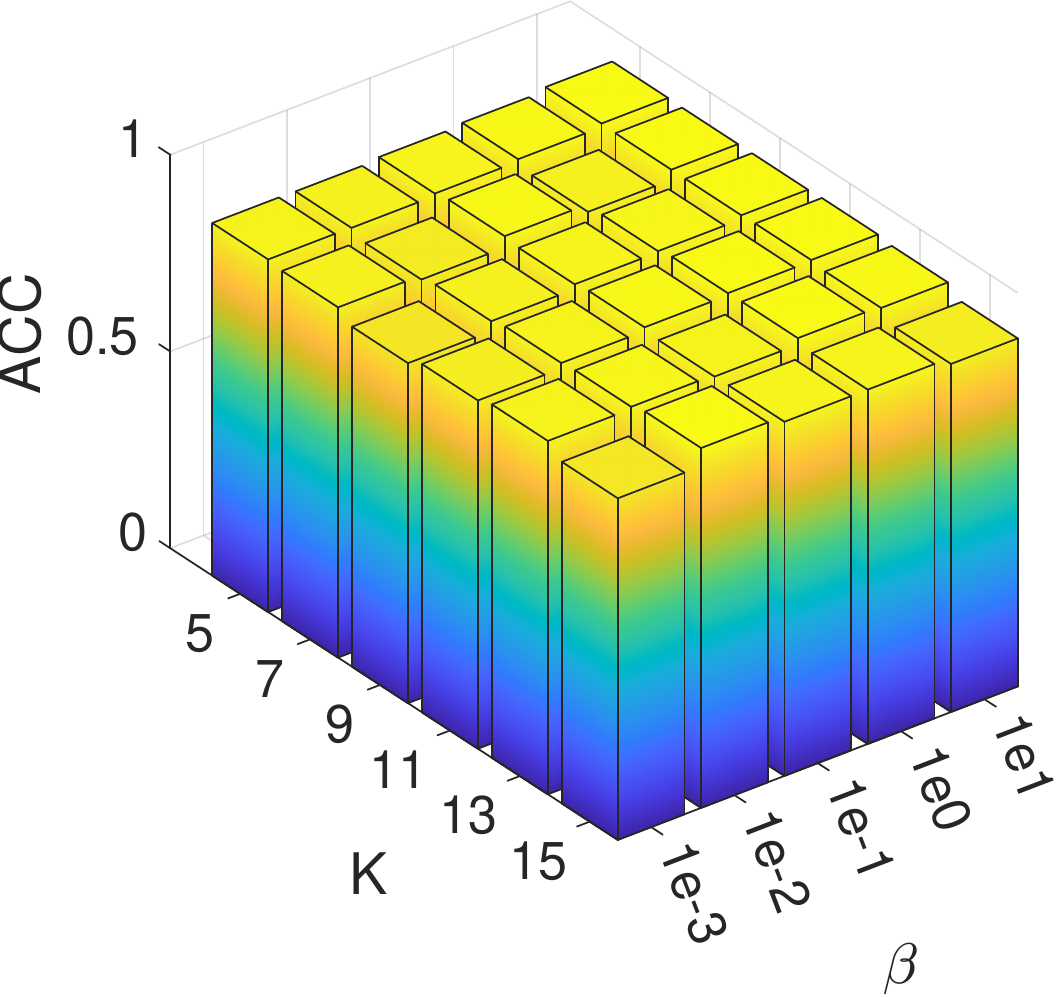}
}
\subfloat[Caltech7]{
	\includegraphics[width=0.49\linewidth]{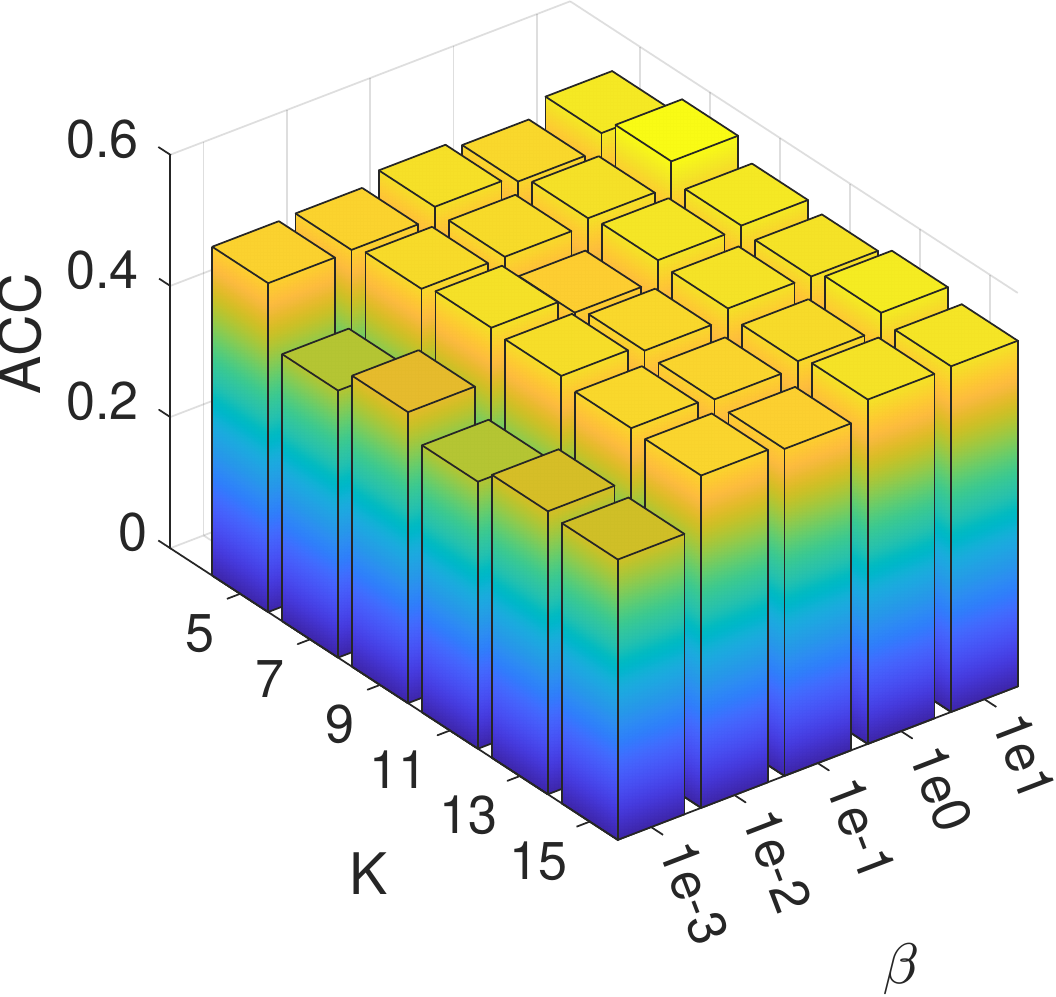}
}
\caption{Results involving different parameter combinations on the Handwritten dataset and Caltech7 dataset with a 50\% missing rate.}
\label{fig.param}
\end{figure}
In our RecFormer, there are two key hyperparameters that need to be set, \textit{i.e.}, $\beta$ and $K$. In order to study the sensitivity of our RecFormer to the two parameters, we perform a mesh search in different parameter combinations and plot the results in Figure \ref{fig.param}. From the figure, it is not difficult to choose a pair of appropriate parameters for our RecFormer. For instance, we can select parameters $\beta$ and $K$ from the range of $[10^{-3},10]$ and $[5,15]$ separately for the Handwritten database, and the range of $[1,10]$ and $[5,15]$ separately for the Caltech7 database.

\section{Conclusion}
In this paper, we present a novel IMC model called RecFormer that uses a two-stage training approach to handle various types of randomly missing data. Distinct from most existing methods, our method focuses on skillfully recovering missing views and performing complete multi-view clustering. To achieve this, we develop a transformer-style cross-view autoencoder to enhance information interaction, and we propose a structure-aware recurrent graph constraint that promotes the restoration of incomplete views and the preservation of geometry structure within the views. This helps to obtain more discriminative semantic fusion information. The key innovation of our model is to leverage known data and the intrinsic geometry structure of data to complete missing views. Sufficient experimental results confirm that our RecFormer has obvious advantages over other top methods. Additionally, our model can be easily extended to other multi-view classification or regression models, providing more robust data support for incomplete multi-view learning by inputting consistent representations into a classifier or regression layer. In future work, we plan to incorporate a measure of completion reliability, enabling the model to selectively fill recovered data into missing locations based on the quality of recovery.



%
\bibliographystyle{IEEEtran}
\bibliography{IEEEabrv}
\vspace{-0.8cm}
\begin{IEEEbiography}[{\includegraphics[width=1in,clip,keepaspectratio]{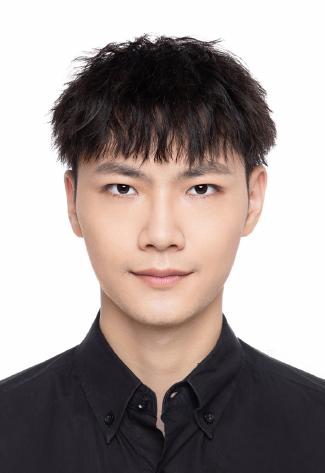}}]{Chengliang Liu}
	(Graduate Student Member, IEEE) received the B.S. degree in computer science from the Jilin University, Changchun, China, in 2018 and the M.S. degree in computer science from the Huazhong University of Science and Technology, Wuhan, China, in 2020. He is currently working toward the Doctoral degree with the Harbin Institute of Technology, Shenzhen, China. His research interests include machine learning and computer vision, especially multiview representation learning.
	\end{IEEEbiography}

	\begin{IEEEbiography}[{\includegraphics[width=1.0in,clip,keepaspectratio]{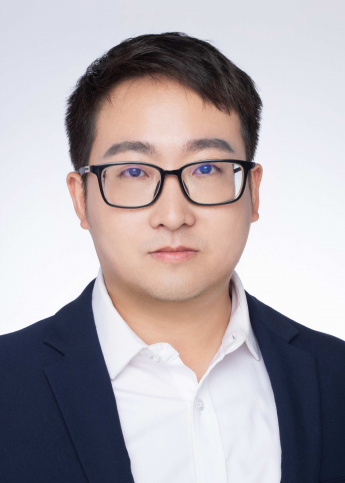}}]{Jie Wen}
	(Member, IEEE) received the Ph.D. degree in Computer Science and Technology, Harbin Institute of Technology (HIT), Shenzhen. He is currently an Assistant Professor at Harbin Institute of Technology, Shenzhen, China. He has published over 50 technical papers at prestigious international journals and conferences, including the IEEE TNNLS, IEEE TIP, IEEE TCYB, IEEE TMM, ECCV, AAAI, IJCAI, ACM MM, etc. His current research interests include, biometrics, pattern recognition and machine learning.
	\end{IEEEbiography}
\vspace{-11pt}
	\begin{IEEEbiography}[{\includegraphics[width=1.0in,clip,keepaspectratio]{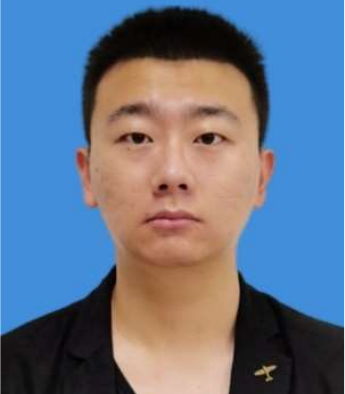}}]{Zhihao~Wu}
	received the B.S. degree in Department of Computer Science from Harbin Engineering University, Harbin, China, in 2019. He is currently pursuing the Ph.D. degree in Harbin Institute of Technology, Shenzhen. His research interests include computer vision and machine learning, especially weakly supervised object detection.
	\end{IEEEbiography}
\vspace{-17pt}
	\begin{IEEEbiography}[{\includegraphics[width=1.0in,clip,keepaspectratio]{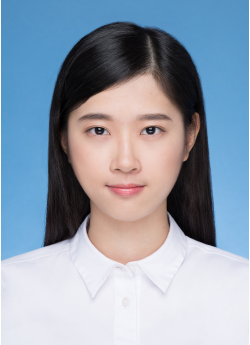}}]{Xiaoling Luo}
		received her B.S. degree in Software Engineering at South Cwhina Normal University (SCNU) in 2017. She is currently pursuing the Ph.D. degree in computer science and technology at Harbin Institute of Technology, Shenzhen, China. Her research interests include pattern recognition, medical image processing and deep learning.
	\end{IEEEbiography}
\vspace{-10pt}
	\begin{IEEEbiography}[{\includegraphics[width=1.0in,clip,keepaspectratio]{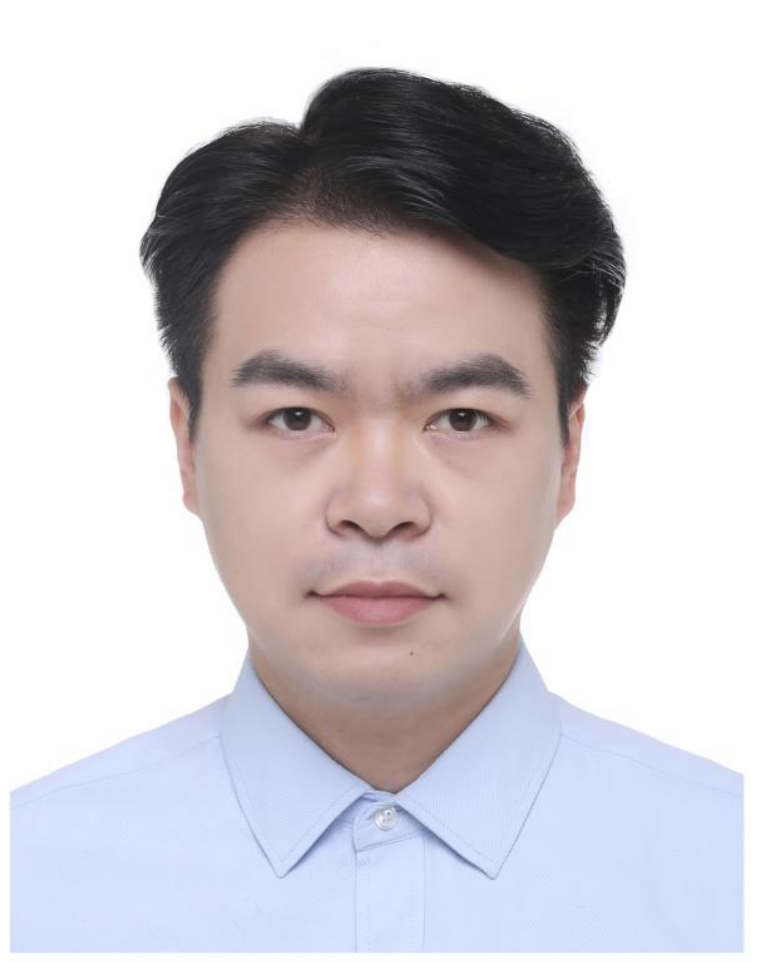}}]{Chao Huang}
	received the Ph.D. degree in computer science and technology from Harbin Institute of Technology, Shenzhen, China, in 2022. From 2019 to 2022, he was a visiting student with Peng Cheng Laboratory. He is currently an Assistant Professor with the School of Cyber Science and Technology, Sun Yat-sen University. His research interests include anomaly detection, multimedia analysis, object detection, image/video compression, and deep learning. Dr. Huang received the Distinguished Paper Award of AAAI 2023, and his dissertation was nominated for Harbin Institute of Technology's Outstanding Dissertation Award. So far, he has published over twenty technical papers at prestigious international journals and conferences. He serves as an Associated Editor for the International Journal of Image and Graphics, and also serves/served as the PC member for several top conferences, including CVPR, ICCV, NuerIPs, ICLR, AAAI, ACM Multimedia, and IEEE ICME. 
	\end{IEEEbiography}
\vspace{-10pt}
	\begin{IEEEbiography}[{\includegraphics[width=1.0in,clip,keepaspectratio]{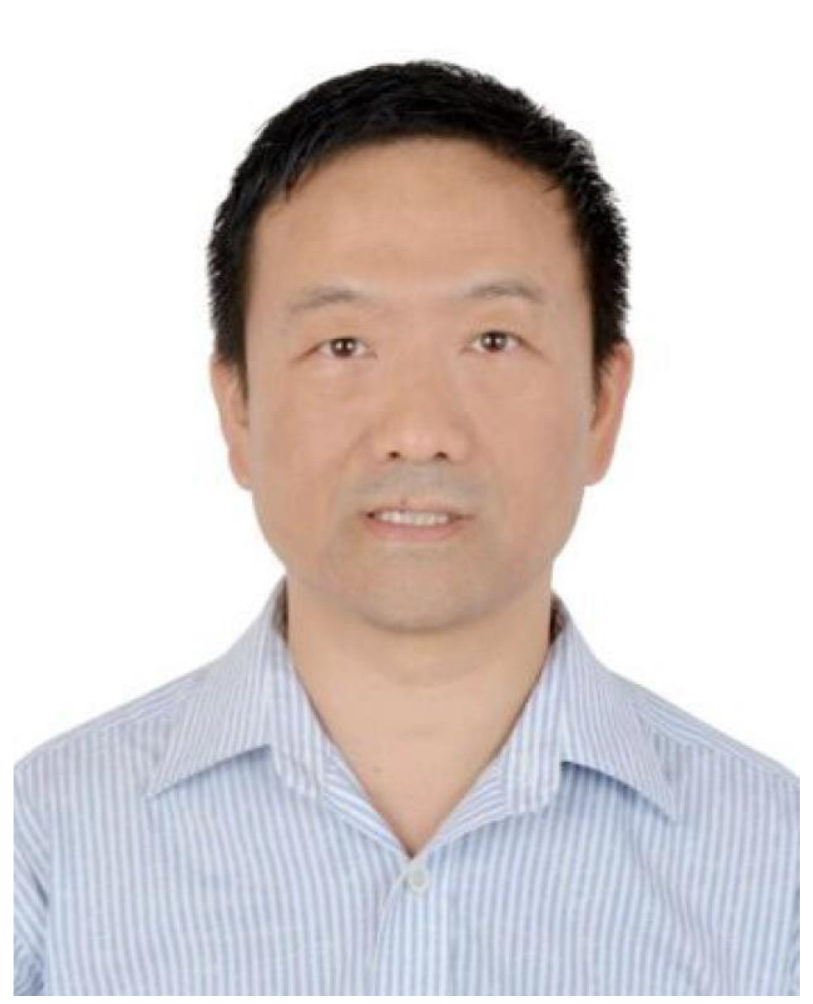}}]{Yong Xu}
	(Senior Member, IEEE) received his B.S. degree, M.S. degree in 1994 and 1997, respectively. He received the Ph.D. degree in Pattern Recognition and Intelligence system at NUST (China) in 2005. He is currently an Professor with the School of Computer Science and Technology, Harbin Institute of Technology (HIT), Shenzhen. His research interests include pattern recognition, deep learning, biometrics, machine learning and video analysis. He has published over 70 papers in toptier academic journals and conferences. His articles have been cited more than 5,800 times in the Web of Science, and 15,000 times in the Google Scholar. He has served as an Co-Editors-in-Chief of the International Journal of Image and Graphics, an Associate Editor of the CAAI Transactions on Intelligence Technology, an editor of the Pattern Recognition and Artificial Intelligence. 
	\end{IEEEbiography}


\vfill

\end{document}